\documentclass[11pt]{article}

\usepackage{etoolbox}
\usepackage{caption}
\usepackage{graphicx,algorithm2e,amsmath,fullpage,booktabs,float,amssymb}   %package of "arydshln" for dashed or dotted lines in table envirement
\DeclareGraphicsExtensions{.pdf,.png,.jpg}
\begin{document}

\title{Ideological Sublations: Resolution of Dialectic in Population-based Optimization}%A Fast and Effective Optimization Algorithm by Dialectical Interactions}

\author{S. Hossein Hosseini and Afshin Ebrahimi\\
ICT Research Lab, Faculty of Electrical Engineering\\ Sahand University of Technology, Tabriz, Iran\\
\{\textit{h\_hosseini, aebrahimi}\}\textit{@sut.ac.ir}}

\providecommand{\keywords}[1]{\textbf{\textit{\small Index terms---}}#1}
%\IEEEoverridecommandlockouts
%\IEEEpubid{}
\date{}
\maketitle

\begin{abstract}

A population-based optimization algorithm was designed, inspired by two main thinking modes in philosophy, both based on dialectic concept and thesis-antithesis paradigm. They impose two different kinds of dialectics. Idealistic and materialistic antitheses are formulated as optimization models. Based on the models, the population is coordinated for dialectical interactions. At the population-based context, the formulated optimization models are reduced to a simple detection problem for each thinker (particle). According to the assigned thinking mode to each thinker and her/his measurements of corresponding dialectic with other candidate particles, they deterministically decide to interact with a thinker in maximum dialectic with their theses. The position of a thinker at maximum dialectic is known as an available antithesis among the existing solutions. The dialectical interactions at each ideological community are distinguished by meaningful distributions of step-sizes for each thinking mode. In fact, the thinking modes are regarded as exploration and exploitation elements of the proposed algorithm. The result is a delicate balance without any requirement for adjustment of step-size coefficients. Main parameter of the proposed algorithm is the number of particles appointed to each thinking modes, or equivalently for each kind of motions. An additional integer parameter is defined to boost the stability of the final algorithm in some particular problems. The proposed algorithm is evaluated by a testbed of 12 single-objective continuous benchmark functions. Moreover, its performance and speed were highlighted in sparse reconstruction and antenna selection problems, at the context of compressed sensing and massive MIMO, respectively. The results indicate fast and efficient performance in comparison with well-known evolutionary algorithms and dedicated state-of-the-art algorithms.

\end{abstract}

%\begin{keyword}


\normalsize
\section{Introduction} \label{Intro}

Optimization is a necessary tool in a lot of fields of science and engineering. Two main approaches of the optimization are based on mathematical methods and metaheuritic ways. Mathematical methods such as gradient-based approaches are reliable with proof of convergence to a global optimum solution under predetermined conditions on the optimization model \cite{BAZ05}. However, the conditions are satisfied only by specific models, and still there are a lot of real-world problems which are not tractable by mathematical optimization. (Meta)heuristics/evolutionary algorithms are appropriate approach at such situations. They have potential to discover the global optimum solution regardless of the properties of cost/fitness function. They operate with minimum information about the function. As a consequence, they are easy to program and adjust for different problems.

Metaheuristic algorithms borrow a kind of intelligence, almost from the nature. They are able to discover a global optimum solution for wide range of problems. Crucial requirement for such intelligence is existence of exploration and exploitation features in the source of inspiration. An intelligent system should be able to exploit a solution confirmed as a promising one, and be able to explore an enough number of candidate solutions, efficiently. As an obvious instance, we can think of natural thinking abilities of a human being. The focused and diffused (default) thinking modes can be regarded as the exploitation and exploration abilities of a mankind, respectively \cite{MHIY07}. They are thinking modes from phycological point of view. Our proposed algorithm is inspired by thinking modes developed at the context of philosophy. During the history, a lot of philosophers developed some thinking modes to equip the mankind with powerful tools at the path of discovering truth \cite{CCC14}. In our terminology at this work, truth is the desired global optimum solution, and the population is equipped with two opposite kinds of thinking modes, i.e., speculative thinking for exploration and practical thinking for exploitation. We borrow the idea of dialectics from the modern philosophy to define the thinking modes and their results. Two types of dialectics are modeled based on distances in objective and subjective spaces. By thinking, we refer to a simple procedure in which each solution (thinker) selects a dialectical solution for interaction.

%\IEEEpubidadjcol

The key in solving different problems is providing a balance between the exploration and exploitation. It is approached by controlling parameters of the algorithm. A perfect balance leads to an efficient search in a reasonable time. Hence, the main point in designing a new algorithm is consistency between the operators developed for exploitation and exploration. It would make the balance to be easily captured by minimum number of parameters. A review on some main operators of a few well-known algorithms gains some insight into the evolution of exploitation and exploration operators. A basic operator for exploitation is the one used in particle swarm optimization (PSO) algorithm \cite{JK95}. The motion of all particles toward best solution has high exploitation power, however, at the expense and risk of being trapped at a local optimum. On the other side, in genetic algorithm (GA) \cite{Holland92}, random mutations driven from a specific probabilistic distribution has significant exploration power, at the expense runtime. From one point of view, other operators introduced in the other metaheuristic algorithms after PSO and GA, try to relax the determinism of the motion toward one leader, and try to constrict the randomness of mutations.

In order to avoid from the local optimums, determinism of the motions toward one leader in PSO was relaxed in its variants. For example, in fitness-distance-ratio based PSO (FDR-PSO) \cite{FDR03}, a near and better solution is selected for each particle to follow a local leader instead of just one global leader. Other algorithms such as imperialistic competition \cite{EA07} and natural aggregation \cite{NNA16}, utilize $k$-best solutions instead of $1$-best leader in PSO. However, there is still randomness in the selection of one of the $k$-best solutions as the imperialist/shelter. On the other side, differential evolution (DE) constricts the completely random mutations of GA by using difference vectors among the solutions as the mutation vectors. However, there is still randomness in selection of two solutions for computation of mutation vector (in DE/rand/1/bin), and also there is a determinism in following one best solution (in DE/best/1/bin) \cite{RS97}. Moreover, same random selections exist in the interactions among solutions of other algorithms such as brainstorm optimization \cite{BS15}, and in learning phase of teaching-learning-based optimization \cite{TLBO11}. Overall, except of PSO and its variants in which all particles follow same criteria, other mentioned algorithms contain a randomness in selection phase of the interactions. Our main motivation in development of the proposed algorithm was discovering a systematic interaction among the particles without any randomness in the selection phase. As explained at next section, the idea was inspired by modern philosophy based on \emph{systematic} dialectic instead of \emph{arbitrary} dialectic utilized among ancient philosophers. At this work, the arbitrary dialectic was interpreted as the result of random selection of particles for interaction. We would develop an efficient deterministic selection scheme by leveraging the definitions of two kinds of antitheses.

In literature, opposite solutions are utilized for acceleration of evolutionary algorithms \cite{SR08}. Further, there is a research direction on high-level language programming inspired by the dialectical philosophy \cite{dia09}. The most related work is an optimization algorithms called dialectic search \cite{SK09}. However, it has fundamental differences with our proposed algorithm at the context of source of inspiration and modeling ways:
\begin{enumerate}
  \item Dialectic search algorithm is a single-solution approach such as simulated annealing \cite{SA83} and tabu search \cite{Tabu89}, while our proposed algorithm is a population-based method.
  \item Dialectic search is inspired by the work of Hegel and Fiche who developed idealistic dialectic, while our source of inspiration is based on both idealistic and materialistic dialectics.
  \item In our models, dialectic is searched among the population, such that the population of solutions improve their positions based on dialectical interactions, while in the dialectic search algorithm, dialectic was imposed by local random changes in the single solution.
  \item In our proposed algorithm, the new solutions are generated by meaningful steps toward the dialectical solutions, known as antithesis, while in the dialectic search a new solution is searched at the path toward the dialectical solution.
\end{enumerate}
It is worth mentioning that our idea of proposed algorithm was formed and developed without being aware of the dialectic search algorithm. The proposed algorithm was named as ideological sublations (IS), because of tendency of the thinkers (particles) for canceling their theses (positions) while they simultaneously preserve it. Two essential ideas behind of IS algorithm are definition of \emph{1)} the Euclidian distance between two solution as a metric of idealistic contradiction and \emph{2)} the difference between their objective functions as a metric of materialistic contradiction. The key for management of the contradictions was separation of the solutions to two groups according to their qualities.

Rest of the paper is organized as follows. At next section, the concept of dialectics and its evolution were reviewed from a philosophical point of view. Also, connections with the proposed algorithm are discussed at this section. At section \ref{proposed}, proposed algorithm was explained after modeling the considered thinking modes. At section 4, experimental results on test benchmark functions, sparse reconstruction problem, and antenna selection challenge were included. Finally, a discussion was provided at section 5, and the paper was concluded at section 6.

\section{A Review on the Evolution of Dialectic} \label{review}

The word of dialect is literally composed of the prefix \textit{dia-} which means "across", and the Greek root \textit{legein} which means "speak" \cite{dic1}. In the context of philosophy, dialectic is a process of contradiction between two opposite sides of everything that leads to truth. First utilization of the dialectic belongs to ancient Greek philosophers who innovated a back-and-forth form of dialectic in their arguments \cite{MJ16}. Later, other dialectical thinking modes were developed and created by different philosophers \cite{CCC14}. In a philosophical expression, the aim was a \emph{universal} thinking mode that eliminates the opposition between thinking and existence in \emph{any situation}. Among the developed modes, two complementary modes of dialectical thinking have got the most attentions; speculative thinking and practical thinking.

Speculative mode of dialectical thinking was radically evolved by G. W. F. Hegel (1770-1831). He reformed the classic version of dialectic. In his systematic model of dialectic, as included in Figure \ref{HD}, speculative moment or the moment of resolution arises after two stages of understanding moment and dialectical (sublation) moment. A thesis that seems stable at the understanding moment, challenges itself (because of its one-sidedness or restrictedness) and pass into its opposite side (antithesis) at the dialectical moment. Contradiction between thesis and antithesis at the unstable moment of sublation, leads to a new emerging and more sophisticated thesis (synthesis) at the speculative moment. At the next repeat, the synthesis challenges itself, interacts with its antithesis, and reforms itself to another new synthesis. The process continues until reaching the truth. In our terminology in the proposed swarm-based optimization algorithm, all candidate solutions are regarded as the existing theses, truth is optimum solution to be discovered, and speculative thinking mode is modeled to explore the search space. Let us introduce the speculative operation as a guess with particular randomness.

Main difference of the Hegelian dialectic with the classical one is the process of \emph{self-sublation} at the dialectical moment. At this process, each thesis cancels out and preserves itself simultaneously, such that it transforms to an antithesis. Hence, despite of classical dialectic that waits for an \emph{arbitrary} opposition from outside, the progress in the Hegel's process is \emph{deterministic} because of the unity of thesis-antithesis in his model. According to Hegel's findings, his procedure leads to an \emph{exact} truth, despite of the ancient method that leads to an \emph{approximate} truth \cite{MJ16}. By this extreme refinement in the definition of dialectic, and expressing the speculative thinking process in the mentioned three logical stages, Hegel introduced a \emph{systematic} idealism in which a systematic and deterministic change in the subjective idea leads to improvement in the objective material. In the context of metaheuristics, if we regard the random mutations of genetic algorithm as the arbitrary dialectics, then differential mutations of DE algorithm follow a more systematic and intelligent way in the production of dialectic. However, still the randomness comes from the arbitrary choice of generating pairs of the mutation vectors (in DE/rand/1/bin). Same kind of randomness exists in the teaching-learning-based optimization algorithm as arbitrary interactions among students. At the proposed algorithm, based on the definition of self-sublation, mutation vectors are generated from two deterministically selected candidate solutions, i.e., idealistic thesis and its available anti-thesis. In fact, speculative thinking mode was used as an exploration operator with eliminated randomness in choosing a pair for each solution vector.

Materialistic dialectic is the complementary part of the idealistic dialectic. It is well-known by K. Marx (1818-1883). On the opposite direction with Hegel's philosophy, Marx refused to speculate in \emph{details} \cite{Marx15}. He realized that the opposition of thinking and existence has root in the human's activities \cite{CCC14}. In the materialistic ideology, a social existence determines consciousness. That was on the contrary with the idealistic thoughts of determination of existence by consciousness. Nevertheless, the idea of materialistic dialectic was also expressed in the same three-logical stages of understanding, dialectic (sublation), and resolution moments which lead to the thesis-antithesis-synthesis paradigm. Practical thinking mode was developed by this kind of dialectic. According to materialism, change in the objective material leads to improvement in the subjective idea. As a symbolic example of such procedure of reformation, we can mention to an important phycological progress of \textit{confidence in the ability to use resources and to master nature}, after the industrial revolution \cite{IndustrialRev16}. Practical thinking mode was translated to population-based optimization context, and utilized as an exploitation operator with a relaxed determinism in selection and movement toward a leader.

\section{Proposed Algorithm} \label{proposed}

\begin{figure}[h]
 \centering
  \includegraphics[scale=0.36]{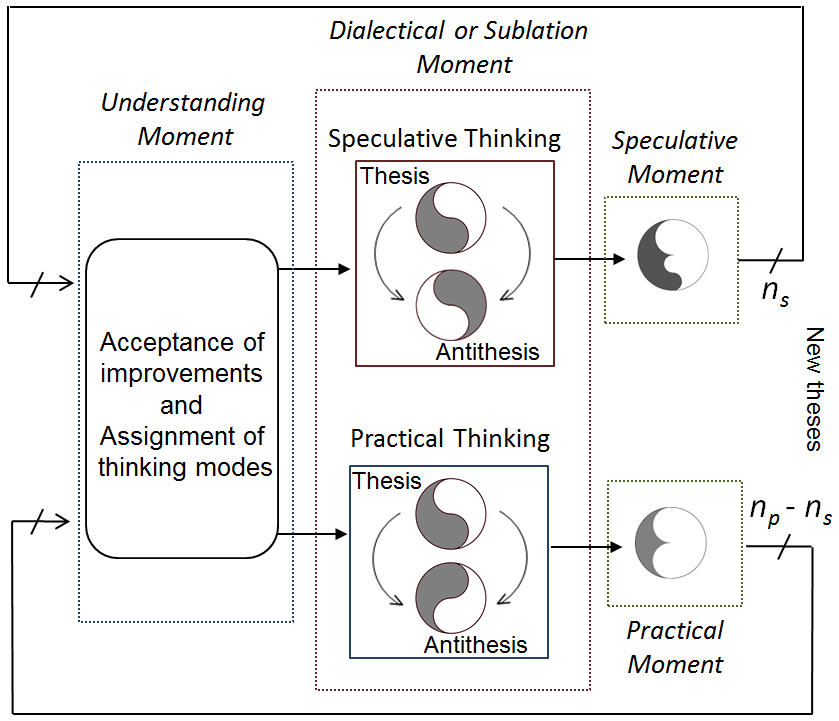}
  \caption{Main Steps of the proposed algorithm expressed in three stages of dialectical logic}
  \label{HD}
\end{figure}

Block diagram in Figure \ref{HD}, illustrates the main idea behind the proposed algorithm. As illustrated, the loop of algorithm consists of three understanding, sublation, and speculative/practical moments. As would be clarified, the understanding and speculative/practical moments are modeled by simple operators regularly utilized in the context of swarm-based optimization algorithms, but with some meaningful nuances. Hence, the operators introduced for sublation moment are the main idea behind of the proposed algorithm. At this section, first, two kinds of difference among the solution vectors in a population were highlighted, consequently two models for the dialectics were formulated. Then, the proposed dialectic models which lead to unique antithesis are translated to the population-based optimization context in the three logical stages.

\subsection{Dialectic Models}

A solution vector $\textbf{x}=[x_{1},~ ...,~ x_{d}]$ is regarded as one individual thesis about $d$ different subjects. The goal is optimizing a cost/fitness function of the decision variables, i.e., $f(\textbf{x})$. In swarm/population based approaches $p$ candidate solutions follow some rules to discover the global optimum solution. Any difference among $p$ theses in the population leads to a challenge for motion. However, as philosophy promises, an extreme difference - called dialectic - can lead to a high-resolution optimum solution (truth), with a higher speed than an arbitrary difference. In order to organize dialectical interactions among the solutions, two idealistic and materialistic antithesis are modeled.

Simply, we define the Euclidian distance between two solutions as the idealistic difference, and the distance in objective space as the materialistic difference. Regardless of limitation on the acquired number of samples from the function, an idealistic antithesis $\textbf{x}^{anti}$ for one specific thesis $\textbf{x}^{thes}$ was modeled as the solution of following optimization problem:

\begin{equation}\label{di1}
  \textbf{x}^{anti}=arg_\textbf{x}\max \|\textbf{x}-\textbf{x}^{thes}\|_2
\end{equation}
\begin{displaymath}
  s.t. ~~ f(\textbf{x})=f(\textbf{x}^{thes}).
\end{displaymath}

According to the proposed model, a thesis whom belongs to the speculative thinking community should sublate itself in such way that leads to an antithesis in largest distance (canceling out property), but at the same level of quality (preserving property). The model (\ref{di1}) is an idealistic definition of speculative antithesis. According to the definition, an exact antithesis is only identifiable, when whole infinite number of the solutions in the domain with the same quality as the thesis $\textbf{x}^{thes}$ are evaluated. Actually, such procedure is not efficient to be practical. However, a mimicked translation of the idea is possible for a community with finite number of population.

On the other hand, a practical antithesis is searchable among a number of best solutions; one of them that is in nearest distance from a practical thesis, has a dialectical position. Since, it is most approachable solution that promises significantly higher qualities. In a mathematical expression:

\begin{equation}\label{di2}
  \textbf{x}^{anti}=arg_\textbf{x}\min \|\textbf{x}-\textbf{x}^{thes}\|_2
\end{equation}
\begin{displaymath}
  s.t. ~~ |f(\textbf{x})-f(\textbf{x}^{thes})|>\Delta,
\end{displaymath}

\noindent where the amount of scalar $\Delta$ guarantees a dialectical gap between a materialistic thesis and its corresponding antithesis. The gap is canceling out a practical thesis in a materialistic self-sublation. On the other side, looking for a closest solution or minimizing the Euclidian distance (idealistic dialectic) is the preserving side of the sublation in practical thinking mode. At the following, we arrange the thinkers/particles in such manner that the desired dialectics are included in the interactions among the thinkers. As indicated, although finding a solution in a perfect dialectic with each candidate solution is not possible, but still approximating an antithesis among the existing solutions (theses) delivers a taste of what dialectical philosophy promises.

\subsection{Understanding Moment}

At this stage all initial/new solutions are evaluated. Except of the first iteration in which all initial solutions are considered as the new theses, at the other iterations a new thesis (synthesis) is only accepted if its position at the current iteration has better quality than its position in previous iteration. On the other expression, the best thesis for each thinker is preserved during the optimization process. Consequently, at each iteration, all accepted solutions as the new theses are sorted according to their cost/fitness values. The sorted theses are divided into two groups of high-quality and low-quality solutions. Simply, $k_1$ high-quality solutions are appointed for speculative thinking and the remaining $p-k_1$ solutions are assigned for practical thinking, where $p$ is total number of thinkers/particles. As elaborated in the next subsection, this sort of assignment simplifies the task of each thinker in finding its corresponding antithesis among all available solutions. The integer value of $k_1$ is the main parameter of IS algorithm that can be easily adjusted by trying some different values between $[2,p-1]$, e.g., multiplication of 5. Large values of $k_1$ provide a high exploration capacity by large number of speculative thinkers, and small values boost the exploitation capability by increasing the number of practical thinkers. As would be seen in simulation results, in most problems an appropriate value is a integer number larger than $k_1=\frac{p}{3}$. As a summary, after each repeat, at the understanding moment, the new theses/positions are checked for acceptance or rejection, and are sorted for assignment of speculative or practical thinking mode to each thinker/particle.

\subsection{Self-Sublation Moment}

\begin{figure}[h] %\footnotesize
 \centering
  \includegraphics[scale=0.32]{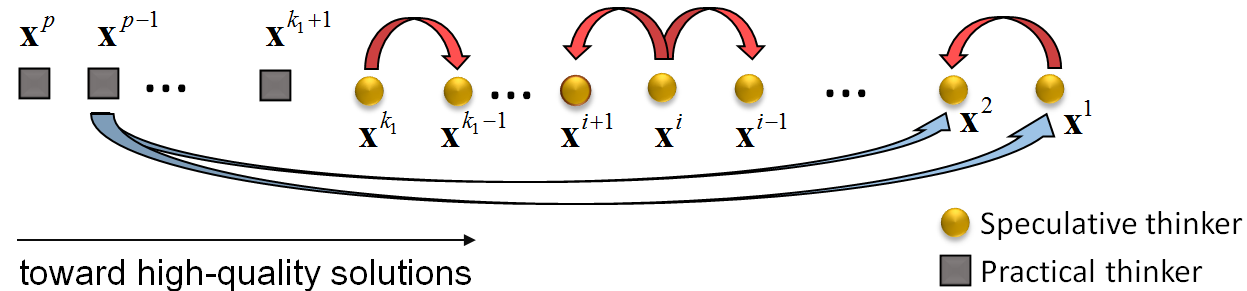}
\caption{Illustration of sublation moment with $k_2=2$; each thinker looks for another candidate thesis which is located in largest distance with similar quality (speculative thinking) or is located in nearest neighbor but among best solutions (practical thinking); a smaller superscript index means a higher-quality thesis.}
  \label{pair}
\end{figure}

At this moment, each thinker challenges her thesis according to the assigned thinking mode at the understanding moment. During this process which is called self-sublation, each thinker looks for an antithesis among the available theses from other thinkers. Discovered antithesis would be a reference point for a change. As mentioned, antithesis is a solution at maximum contradiction or dialectic with the thesis. Of course, since there are two kind of dialectics (idealistic dialectic for speculative thinking and materialistic dialectic for practical thinking), hence, along with the maximization of a particular dialectic, its opposite dialectic is minimized at each thinking mode. This fact is implicitly formulated in the constraint part of model (1), and objective sentence of model (2). At the following subsections, we reduce the proposed models to two simple hypotheses for finding an approximate antithesis for one typical thesis in the collection of solutions. Depending on the assigned thinking mode to each thinker, one of the hypotheses would be deployed. Figure \ref{pair} demonstrates the proposed self-sublation scheme among sorted theses according to their qualities at one specific iteration. At this figure, the indicated solution(s) by arrow are candidate(s) to be considered as an antithesis for their corresponding thesis.

\subsubsection{Speculative Thinking Mode}

The speculative thinking is the mode of thinking among $k_1$ high-quality solutions. Except of first best and $k_1^{th}$ best solutions that deterministically choose their nearest speculative thinker at the objective space as the antithesis, other solutions face with a simple detection problem in finding their antithesis (see Figure \ref{pair}). If we label the speculative solutions by their quality order from $x^1$ for the best solution to $x^{k_1}$ for the $k_1^{th}$ best solution, then the antithesis for first and last theses would be:

\begin{equation}\label{anti1}
\textbf{x}^{anti-i}=
 \begin{cases}
   \textbf{x}^2, & \mbox{if }~ i=1 \\
   \textbf{x}^{k_1-1}, & \mbox{if } ~i=k_1
 \end{cases}
\end{equation}

\noindent Otherwise, for $1<i<k_1$, the remaining speculative thinkers look for antithesis in their objective neighbourhood with radius 1 from themselves. On the other words, each thinker checks the distance of his thesis from the theses of one higher-quality and one lower-quality thinker, then selects one of them who has a thesis in longest distance respect to his thesis. In fact, looking at objective neighborhood is preserving and choosing one solution in largest distance is canceling out a speculative thesis at the self-sublation moment. As another expression, the antithesis for $i^{th}$  thesis with $1<i<k_1$ is:

\begin{equation}\label{anti2}
\textbf{x}^{anti-i}=
 \begin{cases}

   \textbf{x}^{i+1}, & \mbox{if} ~ \|\textbf{x}^{i+1}-\textbf{x}^i\|_2 > \|\textbf{x}^{i-1}-\textbf{x}^i\|_2 \\
   \textbf{x}^{i-1}, & \mbox{if} ~ \|\textbf{x}^{i+1}-\textbf{x}^i\|_2 \leq \|\textbf{x}^{i-1}-\textbf{x}^i\|_2 \\

 \end{cases}
\end{equation}

\noindent Clearly, selection of a solution at large distance as the antithesis, boosts the exploration property of the speculative thinking mode. %That is also worth pointing out that our approach of discovering idealistic antithesis is completely different from the method in \cite{SK09} that imposes random dialectics to one thesis as a tool of self-sublation. Regardless of the limitations, the proposed model in \ref{di1} leads to a unique anti-thesis while the uniqueness was not center of attention in \cite{SK09}. In addition, our definition of idealistic anti-thesis counteracts with an statement in \cite{SK09} which claims that the anti-thesis of a binary vector is its complementary array. According to the proposed model \ref{di1}, this statement in \cite{SK09} is not valid, since there is not any guarantee that the cost function of the complementary vector would be equal or approximately same as the cost function of the original binary vector as the thesis.
\vspace{0.05 in}

\subsubsection{Practical Thinking Mode}

Low-quality solutions - consist of $p-k_1$ thinkers - measure the distance of their theses with the best existing thesis (existing truth) and with its idealistic antithesis. One of them which is closer to that specific practical thesis, is chosen as a practical antithesis. As indicated in equation (\ref{anti1}), the antithesis for best solution ($\textbf{x}^1$) is always fixed on the second best solution ($\textbf{x}^2$). The second best solution is often a reasonable candidate as the antithesis for practical thinkers. However, in some problems, it can leads to stability issues. We define the axillary parameter $k_2$ to increase the stability in such situations. At each iteration, the distance of $k_2$-best solutions (except of best solution) with the best solution are measured, and one solution in largest distance is chosen as an alternative idealistic antithesis $\textbf{x}^{anti-1}$ for best solution $\textbf{x}^1$. This alternative antithesis is an appropriate candidate antithesis for practical thinkers at specific problems, i.e.

\begin{equation}\label{ant3}
\textbf{x}^{anti-1}=arg_{\textbf{x}^i}\max \|\textbf{x}^i-\textbf{x}^1\|_2
\end{equation}
\begin{displaymath}
i=2,~ \ldots,k_2
\end{displaymath}

\noindent the result of practical sublation for $i^{th}$ solution among low-quality solutions ($i=k_1+1,~\ldots,p$) would be as follow:

\begin{equation}\label{anti3}
\textbf{x}^{anti-i}=
 \begin{cases}

   \textbf{x}^{1}, & \mbox{if} ~ \|\textbf{x}^{anti-1}-\textbf{x}^i\|_2 > \|\textbf{x}^{1}-\textbf{x}^i\|_2 \\
   \textbf{x}^{anti-1}, & \mbox{if} ~ \|\textbf{x}^{anti-1}-\textbf{x}^i\|_2 \leq \|\textbf{x}^{1}-\textbf{x}^i\|_2 \\

 \end{cases}
\end{equation}

It is necessary to emphasis that the mentioned alternative idealistic antithesis for the best solution at each iteration, i.e., $\textbf{x}^{anti-1}$ , is just used for practical thinking of $p-k_1$ low-quality solutions, and second best solution $\textbf{x}^2$ is always a fixed antithesis for speculation of the best solution $\textbf{x}^1$. On the other words, when $k_2=2$, the antithesis for $\textbf{x}^1$  from both view point of the practical thinkers and the best thinker are same. This value is recommended number for initial setting of $k_2$. As indicated in the simulations, $k_2=1$ rarely can lead to better performance. At this value, the best thesis is compulsorily regarded as the antithesis for all practical thinkers. Moreover, increasing the amount of $k_2$ to larger values than 2, can lead to stability of the algorithm in optimization of some particular functions. It is worth mentioning that since $k_2\ll k_1$, hence the desired materialistic dialectic is always held. As the final remark, although the antitheses are selected between two similar candidates, but aggregation of such nuanced decisions by the thinkers leads to a significant impact at the final result, after large number of iterations.

\subsection{Speculative and Practical Moment}

At this moment both practical and speculative thinkers update their theses based on their corresponding antithesis. The detected antitheses are used as a reference point for speculative/practical motions. The update rule is simply modeled by the following equation for all thinkers ($i=1,~\ldots,p$):

\begin{equation}\label{mes}
  \textbf{x}^i := \textbf{x}^i+\mu\odot(\textbf{x}^{anti-i}-\textbf{x}^i)
\end{equation}

\noindent where $\mu=[\mu_1,~ ...,~ \mu_j,~ ..., ~\mu_d]$ is a $d$-dimensional vector with random elements as the step-sizes, and $\odot$ indicates an entry-wise multiplication. The main distinguishing point of the speculative/practical motions is distribution of random variables used as the step-size vector $\mu$. The distributions are different and specific for each speculative and practical thinking modes. After checking some basic distributions, we realized that by step-sizes of speculative motions driven from a uniform distribution with a negligible bias, and simultaneously by normal biased distribution for the step-sizes of practical movements, the algorithm always converges , i.e.,

\begin{equation}\label{step}
\mu_j=
 \begin{cases}

   \mathcal{U}(m_1,\sigma_1), & \mbox{for} ~ i=1,~\ldots,k_1 \\
   \mathcal{N}(m_2,\sigma_2), & \mbox{for} ~ i=k_1+1,~\ldots,p \\

 \end{cases}
\end{equation}

\noindent where $j=1, ..., d$. The following parameters were empirically found as appropriate values in dealing with different problems. They are fixed parameters of the proposed algorithm. Two variable parameters, that their adjustment influence in the performance, are the number of speculative thinkers $k_1$ and the number of elites $k_2$ in finding two opposite directions for exploitation. The fixed parameters of the step-sizes are fixed as:

\begin{itemize}
  \item $m_1= 0.0445$

  \item $\sigma_1=1.02$
\vspace{.05 in}
  \item $ m_2=
\begin{cases}
   %0.6, & \mbox{if} ~ k_2=1 \\
   0.6, & \mbox{if} ~ \textbf{x}^{anti-i}=\textbf{x}^1 \\
   0.45, & \mbox{if} ~\textbf{x}^{anti-i}=\textbf{x}^{anti-1} \\
 \end{cases}$
 \\
 \vspace{.05 in}
 for $i=k_1+1, \ldots, p$
  \item $\sigma_2=\begin{cases}

   \sqrt{0.2}, & \mbox{if} ~ k_2=1 \\
   \sqrt{0.5}, & \mbox{if} ~ k_2>1 \\
 \end{cases}$
\end{itemize}

As inferred from the parameters, the mean $m_1$ and standard deviation $\sigma_1$ of the uniform distribution for speculative step-sizes are always fixed on the given values, independent of other parameters or cases. However, the amount of mean $m_2$ for normal distribution of practical step-sizes depends on the materialistic antithesis which is chosen for one specific practical thesis. If best solution $\textbf{x}^1$ was detected as the antithesis, then a bias of 0.6 is imposed to the normal distribution. Otherwise, at the case of selection of $\textbf{x}^{anti-1}$ as the practical antithesis, an smaller bias of 0.45 is utilized because of lower quality of $\textbf{x}^{anti-1}$ respect to the existing truth $\textbf{x}^1$. In a similar interpretation, variance of practical step-sizes is low, when antithesis for all practical theses is the best solution or equivalently $k_2=1$. The reason is priority of exploitation at this case. On the other hand, when $\textbf{x}^{anti-1}$ was also engaged at the practical sublations for $k_2>1$, then diversity is center of attention, hence, in order to boost the diversity, that is logical to release the concentration toward the target antithesis ($\textbf{x}^1$ or $\textbf{x}^{anti-1}$) by increasing the variance of step-sizes.

\begin{figure}[h]
 \centering
  \includegraphics[scale=0.3]{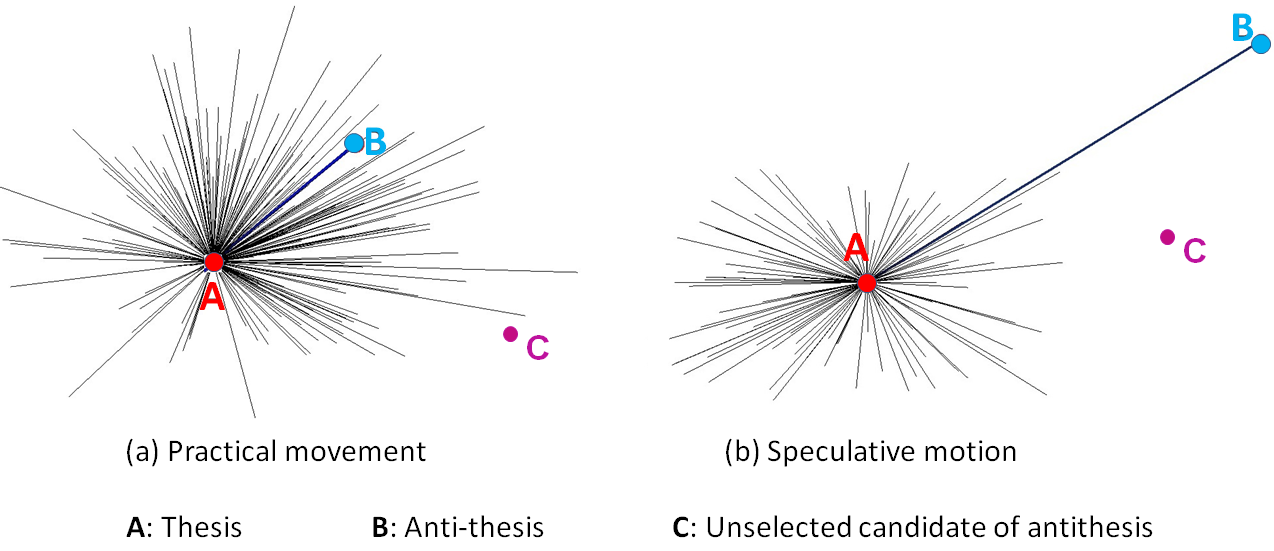}
  \caption{Demonstration of (\textit{a}) attracted steps of practical movements for exploitation and (\textit{b}) naive/uniform search of speculative motions for exploration.}
  \label{distribution}
\end{figure}

Figure \ref{distribution} demonstrates the distribution of random motions for both speculative and practical modes in 2-dimensional space. Start point of the motion is the thesis \textit{A} and the reference point for interaction is detected antithesis \textit{B}. As depicted in Figure \ref{distribution} (\textit{a}), in the practical mode in which significantly better solution(s) are aimed, the thinker scans the area around the antithesis for more \emph{details}. At low variance of step-sizes ($0.2$ for $k_2=1$) the sensing area shrinkages, and becomes more concentrated around the antithesis. Exploitation capability of the algorithm is gained by these practical motions. In comparison with swarm-based global optimization algorithms, such as PSO, FDR-PSO \cite{FDR03}, ICA \cite{EA07}, and NAA \cite{NNA16}, there is a gap between the cost value of the low-quality solutions and their following antitheses. Also, the decision of practical thinkers about their antithesis is deterministically taken between the best solution or its antithesis. That is despite of randomness of ICA and NAA in choosing empire or shelter as a target elite solution.

On the opposite side, the thinkers/particles in speculative mode explore the search space. As illustrated in Figure \ref{distribution} (\textit{b}), in the ideal case, there is not any bias toward the idealistic antithesis, and the thinker with idealistic ideology can freely move in any directions. That is a kind of motion realized by uniform distribution of step-sizes around zero mean. However, in practice, we realized that a little bias of 0.0455 has remarkable impact on the convergence and performance of IS algorithm. In comparison with population-based algorithms, such as genetic and differential evolution, speculative motion can be regarded as a structured mutation which its intensity is controlled by idealistic antithesis. The proposed systematic interactions with deterministic selection of antitheses is in contradiction with the randomness at the selection of generating pairs of mutation vectors in DE algorithm.

%\begin{figure}[h]
% \centering
%  \includegraphics[scale=0.42]{Figs_TA/ProposedAlgorithm/FlowChart.png}
%  \caption{Flowchart of the proposed algorithm}
%  \label{fc}
%\end{figure}

The main stages of the proposed algorithm are summarized at the Algorithm \ref{alg}. Initialization is accomplished in steps 1 and 2. Understanding moment is implemented by steps 3 to 6. Steps 7 and 8 are assigned for sublation and resolution moments, respectively. Obviously, any constraint on the optimization model should be imposed before any function evaluation, i.e., before step 3. As instance, at the case of single objective continuous problems, all decision variables of the candidate solutions ($x_i$s $i=1,\ldots,D$) are preserved at the valid domain of $[x_{min}, x_{max}]$ by passing through two operators of $x_i=min(x_i, x_{max})$ and $x_i=max(x_i, x_{min})$. As another example, in combinatorial problems, the continuous variables should be transformed to the valid discrete variables by an appropriate mapping. At this case, the algorithm usually operates in continuous domain (at step 8), but the evaluations are done at discrete domain. As a remark for combinatorial problems, according to our observations on the antenna selection problem and some other discrete models,

\RestyleAlgo{boxruled}
\LinesNumbered
\begin{algorithm}
\caption{Ideological Sublations (IS) Algorithm \label{alg}}
\emph{Generate a set of $p$ random vectors as initial theses.}\\
\emph{Determine the number of speculative thinkers $k_1$ and elites $k_2$.}\\
\emph{Evaluate all (new) theses.}\\
\emph{Accept a (new) thesis if it leads to improvement in the cost value (except of first iteration).}\\
\emph{Sort the theses according to their cost values.}\\
\emph{Assign first $k_1$ best solutions as the speculative thinkers and rest $p-k_1$ solutions as the practical thinkers.}\\
\emph{Detect the antithesis for each thesis according to the assigned thinking modes to each thinker, using the hypotheses proposed at equations} (\ref{anti1}), (\ref{anti2}), \emph{and} (\ref{anti3}).\\
\emph{Update all theses using their antithesis and assigned step-sizes to each thinking mode, according to equation} (\ref{mes}).\\

\emph{Repeat 3 to 8 until all theses are unified.}
\end{algorithm}

\noindent $l_0$ norm (number of nonzero elements of a vector) is a more appropriate metric for measuring the distances. Replacement of this metric instead of $l_2$ norm (Euclidian distance) at equations (\ref{anti2}), and (\ref{anti3}), gains reasonable decisions about right antitheses, and consequently leads to better performance.

\subsection{On Computational Complexity}

Three main operations determine the order of computational complexity of the proposed algorithm. With $p$ thinkers computational burden at each iteration comes from \emph{1)} sorting of $p$ cost values, \emph{2)} computation of approximately $2p$ distances at the sublation moment, and \emph{3)} computation of $p$ new thesis according to the update rule in (\ref{mes}). Computational complexity of the first operation is $O(p.\log p)$. The second and third operations have similar complexity order of $O(p.d)$. Hence, worst case complexity of IS algorithm after $\frac{nfe}{p}$ iterations is $O(\frac{nfe}{p} . \max (p.\log p~, ~p.d))$, where $nfe$ indicates the number of function evaluations. As a result, the asymptotic order of complexity remains $O(d.nfe))$, since $d > \log p$. Although, the computational complexity of IS algorithm is the same order of magnitude as that of DE \cite{NeiMutDE09}, and roughly at the same order as most of the evolutionary algorithms, but in a fine comparison, as shown in the simulation results, runtime of the operators in the proposed algorithm is at least half of the other test algorithms.

\section{Simulation Results} \label{simulation}

At this section, we evaluate the efficiency and speed of the proposed algorithm using a number of benchmark single objective cost functions, a continuous optimization model for sparse reconstruction, and a binary optimization problem for antenna selection in large scale. For benchmark functions, comparisons were obtained with DE/rand/1/bin (DE for brevity), cooperative DE (CoDE) \cite{CoDE11}, comprehensive learning PSO (CLPSO) \cite{CLPSO06}, grey wolf optimization (GWO) \cite{GWO14}, and teaching-learning-based optimization (TLBO) \cite{TLBO11}. For sparse reconstruction problem, additional comparisons were provided with the PSO by constriction coefficients (PSO-cc) \cite{Kennedy02}, and also with the state-of-the-art dedicated algorithms for sparse reconstruction. Finally, in antenna selection problem, comparisons are provided with the only competitive algorithm among the considered algorithms, and a GA-based discrete algorithm recently proposed for antenna selection. We should mention that our previously developed algorithm - inspired by tornado's air currents \cite{STO16} - despite of its efficiency in low-dimension problems, quickly failed to be competitive in large scales.

Performance metric was cost/fitness value for all problems expect of $f_{12}$ and the sparse reconstruction model, where distortion from optimum solution was also measured. Two metrics of distortion were utilized; mean squared error (MSE) $E[\|\mathbf{x}^\ast-\hat{\mathbf{x}}\|_2]$, and normalized mean squared error (NMSE) $E[{\frac{\|\mathbf{x}^\ast-\hat{\mathbf{x}}\|_2}{\|\mathbf{x}^\ast\|_2}}]$, where $\mathbf{x}^\ast$ is the optimum solution, $\hat{x}$ is the approximated solution, and the expectation operator $E(.)$ indicates averaging over a number of trials. One trial of an optimization procedure was regarded as a successful optimization, if the approached cost value was less than a threshold value of $tr$. The number of thinkers/particles or population size was fixed on 40 for all algorithms, otherwise it is mentioned. All simulations are run on the same computer with Intel Core i3-1.9GHz and 4GHz of RAM operating on Windows 8, 64 bit and MATLAB 2008.

\subsection{Benchmark Functions}

\begin{figure}[h]
 \centering
  \includegraphics[scale=0.21]{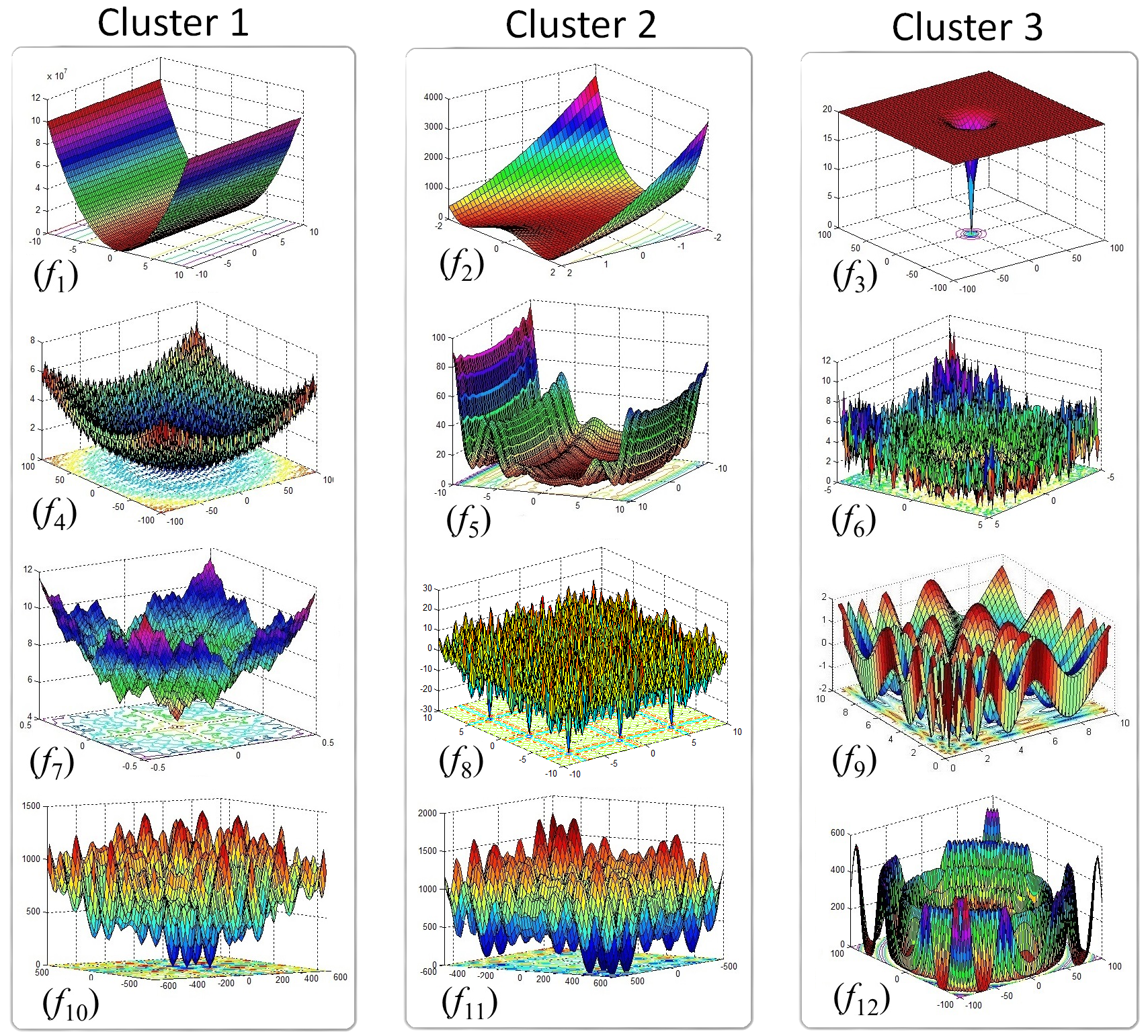}
  \caption{Illustration of benchmark cost functions in two dimensions}
  \label{func}
\end{figure}

At this subsection, 12 benchmark cost functions were used for the evaluation. They were selected among challenging problems of CEC 2017 competition \cite{bench17} and \cite{MJ13}. The considered set of benchmark functions consists of unimodal/multimodal and (non)differentiable functions with (non)separable decision variables. The test functions with 2 variables are depicted in Figure \ref{func}. As shown in the figure, the functions are clustered in such way that in each group specific test algorithm(s) outperforms other ones. The functions and their details are summarized in Table \ref{tab1}. The optimum cost value for all problems is zero, except of three functions of $f_7-f_9$ that their minimum cost values are dependent to the dimension of problem. Moreover, all functions were tested on both 10 and 100 dimensions, expect of last three ones ($f_{10}-f_{12}$) which are only optimized in 40, 80, and 10 dimensions, respectively.

\begin{table}[h!]\scriptsize
 \caption{Definitions of single objective benchmark cost functions}
\begin{center}
  \begin{tabular}[scale=0.2]{p{1.5cm} l r r}
 \hline
 \noalign{\vskip 0.1cm}
   Function &  Formulation &  Domain   \\
  \hline
 \noalign{\vskip 0.2cm}
  Cigar & $f_1=x_1^2+10^6\sum_{i=2}^{d}x_i^2$ & $[-100,100]^d$ \\
\\

  Rosenbrock & $f_2=\sum_{i=1}^{d-1}(100(x_{i+1}-x_i^2)^2+(x_i-1)^2)$ & $[-30,30]^d$ \\
\\

  Easom & $f_{3}=-20\exp(-0.2\sqrt{\frac{1}{d}\sum_{i=1}^{d}{x_i^2}})-\exp(\frac{1}{d}\sum_{i=1}^{d}\cos(2\pi x_i))+20+e$
  & $[-100,100]^d$ \\
\\

 Griewank & $f_4=\sum_{i=1}^{d}\frac{x_i^2}{4000}-\prod_{i=1}^{d}\cos(\frac{x_i}{\sqrt{i}})+1$ & $[0.25,10]^d$\\
\\

 \vspace{.05 in}
  Levy & $f_5=\sin^2(\pi w_1)+1\sum_{i=1}^{d-1}(w_i-1)^2[1+10\sin^2(\pi w_i+1)]+(w_d-1)^2[1+\sin^2(2\pi w_d)]$ & $[-10,10]^d$  \\

  $~$&$~~~~~~~~w_i=1+\frac{x_i-1}{4}, ~ \forall  ~i=1,\ldots~,d$ &\\
\\

 Stochastic & $f_6=\sum_{i=1}^{d}\epsilon_i |x_i-\frac{1}{i}|$ & $[-5,5]^d$ \\
\\

\vspace{.05 in}
 Weierstrass & $f_7=\sum_{i=1}^{d}[\sum_{k=0}^{kmax}a^k\cos (2\pi b^k(x_i+0.5))-d\sum_{k=0}^{kmax}a^k \cos (\pi b^k)]$ & $[-0.5,0.5]^d$  \\

   &$~~~~~~~~a=0.5,~ b=3,~ kmax=20$&\\
\\

 Shubert 3& $f_8=\sum_{i=1}^{d}\sum_{j=1}^{5}j\sin [(j+1)x_i]+j$ & $[-10,10]^d$  \\
\\

 Vincent & $f_9=-\sum_{i=1}^{d}\sin (10\log(x_i))$ & $[0.25,10]^d$ \\
\\

\vspace{.05 in}
 Modified Schwefel & $f_{11}(x)=418.9829\times d-\sum_{i=1}^{d}g(z_i)~~~~~z_i=x_i+4.209687462275036\times 10^2$ & $[-600,600]^d$ \\
&
$g(z_i)=
 \begin{cases}
   z_i\sin(|z_i|^{0.5}), & \mbox{if } |z_i|\leq 500 \\
   (500-mod(z_i,500))\sin(\sqrt{|500-mod(z_i,500)|})-\frac{(z_i-500)^2}{10^4\times d}, & \mbox{if } z_i>500 \\
   (mod(|z_i|,500)-500)\sin(\sqrt{|mod(|z_i|,500)-500|})-\frac{(z_i+500)^2}{10^4\times d}, & \mbox{if } z_i<-500
 \end{cases}$

\\
\\
 -- & $f_{12}=f_{11}$ ~~~ $\forall ~z_i=x_i+4.209687462275036\times \exp(2)$ &$[-600,600]^d$ \\
\\

 Schaffer 7& $f_{10}=\sum_{i=1}^{d}\sum_{j=1}^{5}j\sin [(j+1)x_i]+j$ & $[-100,100]^d$ \\[0.2cm]

 \hline

  \label{tab1}

\end{tabular}
\end{center}
\end{table}

Two parameters of the proposed algorithm were adjusted at each problem for a fast and smooth convergence. Adjusted parameter values were fixed for both small and large dimensions of each problem. Also, the parameters of DE algorithm were carefully tuned for a fair competition with the proposed algorithm. Other algorithms were implemented in their original parameter-free version (TLBO) or by the recommended relations for their parameters (GWO). Most of the variants of DE algorithm were developed with the aim of getting ride of the parameter tuning task in facing different problems. Generally, in one specific application, tuning of an original variant - popularly DE/rand/1/bin - is preferred. An overview of literature on the applications of DE algorithm proofs this statement. However, in literatures, there is a lack of comparison between tuned-DE algorithm and its variants. Here, one of the popular variants of DE, i.e., CoDE, was included in the simulations to justify the reason behind of popularity of original variant of DE in one specific application/problem. On the other hand, most of the variants of PSO try to avoid from trapping in the local optimums. The CLPSO as a well-known variant was used in the comparisons, as well. Available source codes of the comparative algorithms, were utilized in the simulations \cite{codeYarpiz},\cite{codeGWO},\cite{codeCoDE},\cite{codeCL}.

Table \ref{tab2} summarizes the parameter values of IS and DE algorithms. As inferred, an appropriate value for the parameter $k_2$ is usually 2. This value is also an effective choice for sparse reconstruction problem. Larger integer numbers for $k_2$ were used in the functions $f_2$ and $f_5$, in order to increase the stability or reduce the sensitivity to initial solutions. Moreover, smaller integer, i.e., $k_2=1$ was applied to $f_3$, in order to have a special exploitation property. In addition, for most of the problems, an effective integer number for $k_1$ is larger than half of the population size $p$. As shown in the table, the integers assigned to the number of speculative thinkers $k_1$ are almost multiplications of 5. Hence, sensitivity to the parameter $k_1$ is low, and a fine adjustment is not required. An appropriate integer for $k_1$ can be easily found by trying some limited number of possibilities, while $k_2$ is fixed on 2. After adjustment of $k_1$, the parameter $k_2$ can be increased if there was a satiability issue in different runs, or can be decreased if a specific exploitation property is desired.

\begin{figure*}[h!]
 \centering
  \includegraphics[scale=0.205]{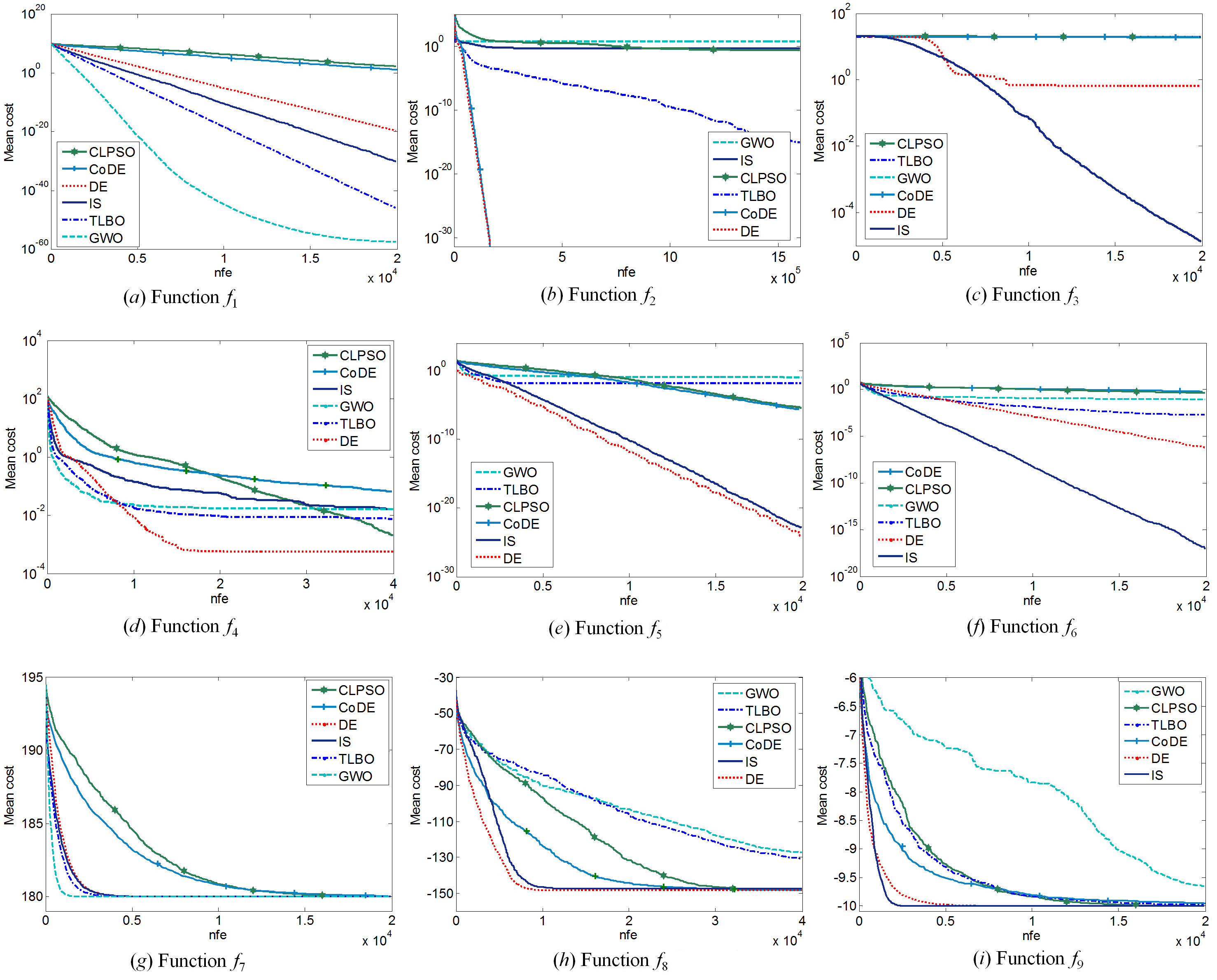}
  \caption{Convergence curves of $f_1-f_9$ in 10 dimensions.}
  \label{ss}
\end{figure*}

\begin{figure*}[h!]
 \centering
  \includegraphics[scale=0.205]{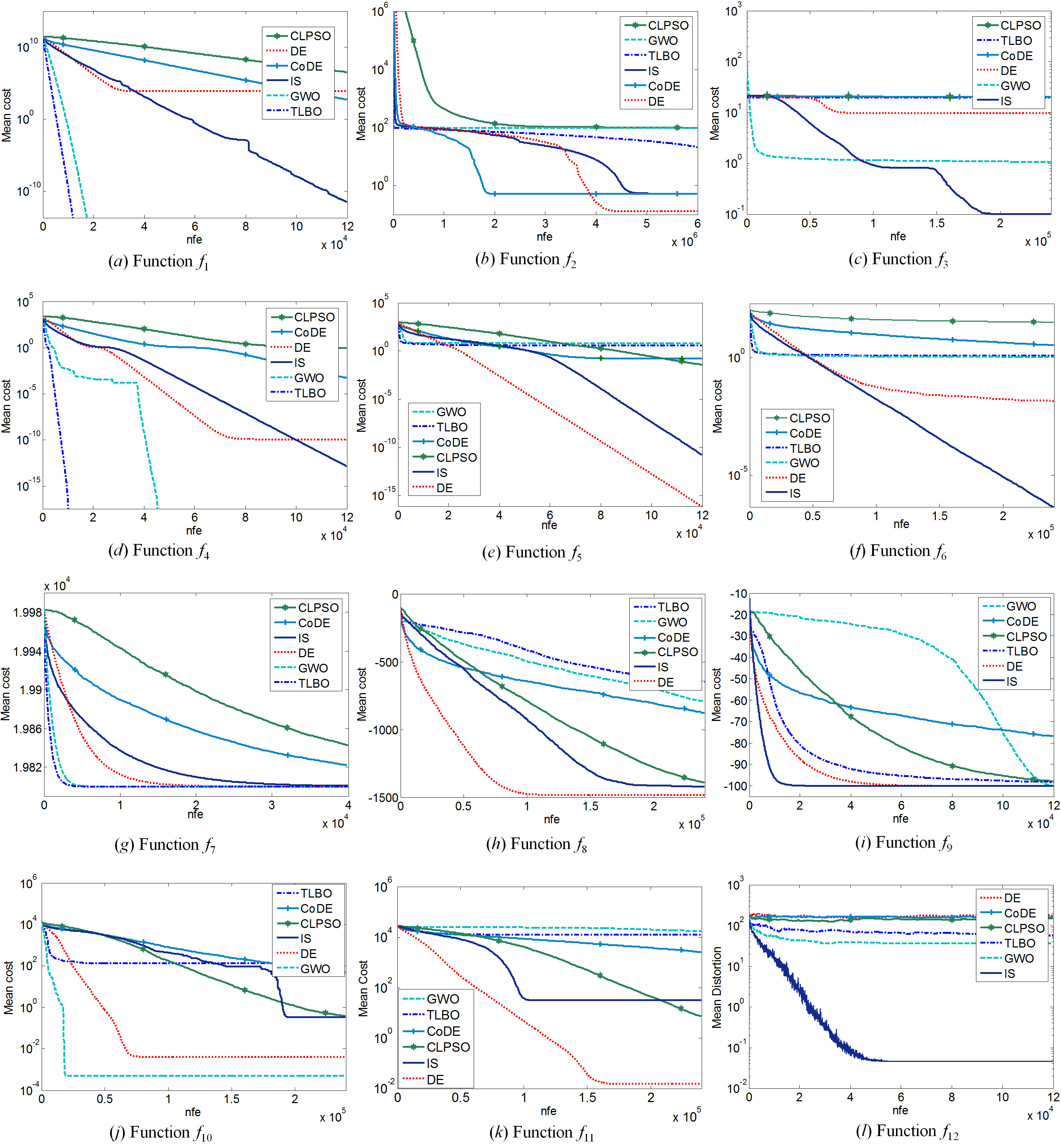}
  \caption{Convergence curves of $f_1$ to $f_9$ in 100 dimensions, and $f_{10}$ to $f_{12}$ in 40, 80, and 10 dimensions, respectively.}
  \label{ls}
\end{figure*}

All algorithms were initiated with same initial solutions, and all of them were stopped after a predetermined number of function evaluations ($nfe$). The initial solutions were produced randomly with the values distributed uniformly within the predefined domains. The domains, and $nfe$ for each scale of the problems were listed in Table \ref{tab1} and Table \ref{tab3}, respectively. Figure \ref{ss} and Figure \ref{ls} show the convergence curves for the functions $f_1$ to $f_9$ in 10 and 100 dimensions, respectively. Also, the convergence curve of the three last benchmark functions ($f_{10}$-$f_{12}$) were included in Figure \ref{ls}. The curves were obtained by averaging over 30 independent runs for each problem. Mean and standard deviation of the best cost value at the final iteration of each problem were reported in Table \ref{tab3}. The mean values smaller than $10^{-20}$ were shown by zero. Clustering of the benchmark functions was based on the number of competitive solutions with the global optimum solution or equivalently the number of global minimums, and also based on the regularity in allocation of the local minimums. This division conduces a rough conclusion about the possible functions in which the proposed algorithm hopefully has better performance.

\begin{table}[h!] \small
 \caption{Settings for DE and IS}
\begin{center}
  \begin{tabular}{l  c  c  c c  c  }
 \hline
 \noalign{\vskip 0.1cm}
         & \multicolumn{2}{c}{DE} & & \multicolumn{2}{c}{IS}\\

           & $Cr$ & $F$ &~& $k_1$ & $k_2$ \\
      \hline
 \noalign{\vskip 0.2cm}
\vspace{.01 in}
    $f_1$ & 0.2 & 0.3 &~& 25 & 2 \\
   \vspace{.01 in}

    $f_2$ & 0.7 & 0.6 &~& 25 & 4 \\
   \vspace{.01 in}

    $f_3$ & 0  & 0.5 &~& 39 & 1  \\
   \vspace{.01 in}

    $f_4$  & 0.1 & 0.3 &~& 30 & 2 \\
   \vspace{.01 in}

    $f_5$ & 0.1  & 0.4 &~& 35 & 12 \\
    \vspace{.01 in}

    $f_6$ & 0.5  & 0.3 &~& 15 & 2 \\
   \vspace{.01 in}

    $f_7$ & 0.2  & 0.2 &~& 30 & 2 \\
   \vspace{.01 in}

    $f_8$  & 0 & 0.4 &~& 25 & 2 \\
   \vspace{.01 in}

    $f_9$ & 0.3  & 0.2 &~& 15 & 2 \\
   \vspace{.01 in}

    $f_{10}$ & 0.01  & 0.9 &~& 35 & 2 \\
    \vspace{.01 in}

    $f_{11}$ & 0.01  & 0.9 &~& 35 & 2 \\
    \vspace{.05 in}

    $f_{12}$ & 0.1  & 1.2 &~& 35 & 2 \\

 \hline

  \label{tab2}

\end{tabular}\\
\end{center}
\end{table}

Cluster 1 consists of prototype examples such as unimodal function without the competitive solutions ($f_1$), multimodal function with regular allocation of non-competitive local minimums ($f_4$), multimodal function with negligible irregularity in the allocation of local minimums, but still without serious competitive solution(s) ($f_7$). Finally, as a most challenging problem at this cluster, there is $f_{10}$ with similar structure as the $f_7$ but with existence of competitive solutions located in a near distance from the global optimum. In general, IS algorithm successfully optimizes the functions at this cluster. However, there are some failures in solving small-scale version of the $f_4$, and also in discovering of global optimal solution of $f_{10}$ (see Figure \ref{ss}(\textit{d}) and Figure \ref{ls}(\textit{j})).

According to the Table \ref{tab3}, large standard deviation of IS algorithm at solving the 10-dimension function of $f_4$ indicates an unstable optimization by this algorithm. As indicated in Table \ref{tab4}, the number of successful optimization by IS algorithm for this problem is 24 from total number of 30 trials. It is highest success rate after 27 exact approximations of DE algorithm. It is worth mentioning that the similar structure of the function $f_4$ (regular positioning of the local minimums such that there is explicit routes toward global optimum) exists in the \textit{Rastrigin} and \textit{Ackley} functions \cite{MJ13}. According to our observations, IS algorithm was also unstable in low dimensions of these problems, and had poor performance in large dimensions. That is despite of its performance in large dimension of $f_4$ that was stably optimized by IS algorithm. On the other hand, small standard deviation of the proposed algorithm in solving the problem $f_{10}$, implicitly indicates that IS algorithm always discovers a competitive local optimal solution of this function. However, according to our observations, by shrinking the domain of decision variables, the global optimum solution is approachable by IS. Roughly speaking, GWO algorithm along with TLBO (in most cases) are best algorithms for optimization of the functions with similar properties as the functions in cluster 1.

\begin{table}[h]\small
 \caption{Statistical results of optimization of benchmark functions; Mean (standard deviation)}
\begin{center}
  \begin{tabular}{l l l l  l  l l l r}
 \hline
 \noalign{\vskip 0.1cm}
    & D & nfe & DE & CoDE & CLPSO & TLBO & GWO & IS\\
      \hline
 \noalign{\vskip 0.2cm}
    $f_1$  & 10 & 20000  &  2.10e-20 & 19.815 & 217.029 & 0 & \textbf{0} &  0 \\
   \vspace{.02 in}
           &    &        &  (2.41e-20) & (15.322) & (197.997) & 0 & 0 &  (1.94e-30)\\

           & 100 & 120000  &  7646.5 & 530.755 & 3.3832e+6 & \textbf{0} & 0 & 3.491e-12 \\
      \vspace{.07 in}
           &    &        &  (32529) & (228.913) & (5.3458e+5)& (0) & (0) & (6.383e-12)  \\

    $f_2$  & 10 & 1.6e+6  &  \textbf{0} & 0 & 0.226 &  7.81e-16 & 6.085 & 0.531 \\
  \vspace{.02 in}
           &    &        &  (0) & 0 & (0.335) & (4.27e-15) & (0.783) &  (1.378)\\

           & 100 & 6e+6  &  \textbf{0.1329} &   0.5315 & 96.241 & 21.293 & 96.645 &  0.5316\\
     \vspace{.07 in}
           &    &        &  (0.7279) & (1.5219)&  (22.401) & (7.578) & (1.126) & (1.378)  \\

    $f_3$  & 10 & 20000  &   0.6655 & 18.872 & 18.891 & 20.00 & 20.221 & \textbf{1.35e-5} \\
  \vspace{.02 in}
           &    &     &  (3.6449) & (0.972) & (1.976) & (1.74e-8) & (0.105) & (1.14e-5) \\

           & 100 & 240000  &  9.684 & 20.283& 20.181 & 20.00 & 21.205 &  \textbf{0.1003} \\
      \vspace{.07 in}
           &    &        &  (8.755) & (0.0253) & (0.0186) & (9.242e-10) & (0.0273) & (0.3064) \\

     $f_4$  & 10 & 40000  &  \textbf{5.67e-4} & 0.0671 & 2.1e-3 & 7.6e-3 & 0.0163 &  0.0164 \\
   \vspace{.02 in}
           &    &        &  (1.9e-3) & (0.0158) & (2.7e-3) & (0.0101) & (0.0196) & (0.0390) \\

           & 100 & 120000  &  1.198e-10 & 5.598e-4 & 0.9072 & \textbf{0} & 0 & 1.505e-13 \\
      \vspace{.07 in}
           &    &        &  (6.078e-10) & (2.1e-3) & (0.0506) & (0) & (0) & (1.321e-13) \\

   $f_5$  & 10 & 20000  &  \textbf{1.12e-25} &  2.49e-6 & 3.64e-6 & 0.0149 & 0.1129 &  1.27e-23\\
   \vspace{.02 in}
           &    &        &  (9.92e-26) & (2.04e-6) & (2.78e-6) & (0.0413) & (0.0965) & (3.17e-23) \\

           & 100 & 120000  & \textbf{7.933e-17}  & 0.1632 & 0.0392 & 3.9043 & 6.2476 &  1.683e-11 \\
      \vspace{.07 in}
           &    &        & (2.206e-17) & (0.3394) & (8.6e-3)  & (0.4688) & (0.4753) & (3.375e-11) \\

    $f_6$  & 10 & 20000  &  6.38e-7 & 0.6115 & 0.4328 & 2.0e-3 &  0.0870 & \textbf{8.82e-18}  \\
   \vspace{.02 in}
           &    &        &  (7.70e-7) & (0.2452) & (0.1118) & (7.5e-3) & (0.0904) & (4.82e-17)  \\

           & 100 & 240000  &   0.0147 & 3.458 & 31.997 & 1.2187 &  1.0726 & \textbf{4.426e-7}
 \\
      \vspace{.07 in}
           &    &        &  (0.0474) & (1.2254) & (2.135) & (0.0915) &  (0.1249) & (3.541e-7)  \\

     $f_7$  & 10 & 20000  &  179.9999 & 180.0485 & 180.0180 & 179.9999 & \textbf{179.9999} & 179.9999 \\
   \vspace{.02 in}
           &    &        &  (0) & (9.9e-3) & (3.8e-3) & (0) & (0) & (0) \\

           & 100 & 40000  &  19800 & 19822 & 19843 & \textbf{19800} & 19800 & 19801 \\
      \vspace{.07 in}
           &    &      &  (0.1684) & (1.634) & (1.9189) & (0) & (3.510e-12) & (0.2165) \\

     $f_8$  & 10 & 40000 &  \textbf{-148.379} & -148.175 & -148.227 & -130.486 & -127.157 & -147.293  \\
   \vspace{.02 in}
           &  &  &  (5.01e-28) & (0.0897) & (0.0896) & (11.9264) & (12.123) & (4.2954) \\

           & 100 & 20000  &  \textbf{-1483.8} & -878.006 & -1392.4 & -645.545 & -791.886 & -1423.3 \\
     \vspace{.07 in}
           &    &        &  (3.856e-7) & (30.891) & (12.227) & (101.377) & (53.282) & (35.401) \\

    $f_9$  & 10 & 20000  &  -10 & -9.9503 & -9.9964 & -9.9732 & -9.6482 & \textbf{-10} \\
   \vspace{.02 in}
           &    &        &  (6.32e-12) & (0.0172) & (2.9e-3) & (0.0710) & (0.7657) & (0) \\

           & 100 & 120000  &  -99.9838 & -76.806 & -97.8412 & -98.1905 & -99.9783 & \textbf{-100} \\
      \vspace{.07 in}
           &    &        &  (0.0219) & (1.5443) & (0.2770) & (2.6162) & (4.6e-3) & (2.39e-10) \\

 $f_{10}$  & 40 & 240000  &  4.2e-3 & 49.6217 & 0.3765 & 135.595 &\textbf{ 5.09e-4} & 0.3305 \\
   \vspace{.07 in}
    &  &  &  (7.9e-3) & (12.6485) & (0.0681) & (412.933) & (1.56e-12) & (0.0539)  \\

 $f_{11}$  & 80 & 240000  &  \textbf{0.0151} & 2537.6 & 7.2362 & 12268 & 17738& 31.2504 \\
   \vspace{.07 in}
           &    &        &  (9.7e-3) & (484.028) & (1.5464) & (1536.2) & (1422.4) & (59.4805) \\

 $f_{12}$  & 10 & 120000  &  1.3534e-4 & 4.166e-04 & 8.108e-5 & 0.0195 & 5.7e-3 & \textbf{1.525e-8} \\
   \vspace{.07 in}
           &    &        &  (1.275e-4) & (3.857e-4) & (9.950e-5) & (0.0223) & (0.0127) & (5.698e-8)  \\

 \hline

  \label{tab3}

\end{tabular}
\end{center}
\end{table}

\begin{table}[h]\small
 \caption{Number of successful optimizations that lead to a cost value less than threshold $tr$ (among 30 trials)}
\begin{center}
  \begin{tabular}{l l l c  c  c c c l}
 \hline
 \noalign{\vskip 0.1cm}
 % & D  & VTR & \multicolumn{3}{|c|}{DE} & \multicolumn{3}{|c|}{TLBO}& \multicolumn{3}{|c} {IS}\\

    & $D$ & $tr$ & DE & CoDE & CLPSO & TLBO & GWO & IS \\

      \hline
 \noalign{\vskip 0.2cm}
\vspace{.05 in}
    $f_1$   & 100 & e-3  &  21 & 0 & 0 & 30 & 30 & 30\\

 \vspace{.02 in}
    $f_2$  & 10 & e-3  &  30 & 30 & 0 &  30 & 0 & 26 \\
  \vspace{.05 in}
           & 100 & e-3  &  29 &   24 & 0 & 0 & 0 &  26\\

 \vspace{.02 in}
    $f_3$  & 10 & e-3  &  29 & 0 & 0 &  0 & 0 & 30 \\
    \vspace{.05 in}
            & 100 & e-3  & 13 & 0 & 0 & 0 & 0 &  27 \\

  \vspace{.05 in}
     $f_4$  & 10 & e-3  &  27 & 0 & 15 & 16 & 13 &  24 \\

  \vspace{.05 in}
    $f_6$  & 10 & e-3  &  23 & 0 & 0 & 0 & 0 &  30 \\

\vspace{.05 in}
 $f_{11}$  & 80 & 1 &  30 & 0 & 0 & 0 & 0 & 23 \\

 \hline

  \label{tab4}

\end{tabular}\\
\end{center}
\end{table}

For functions in cluster 2, the competitive optimal solutions exist in relatively large distance respect together, or the local minimums are arranged in irregular positions. As instance, function $f_{11}$ - which is a shifted-variable version of $f_{10}$ - has local minimums distributed in different positions over its domain. At the functions $f_5$ and $f_8$, in both small and large scales, the proposed algorithm competes with the DE algorithm as the most appropriate algorithm for this cluster. As mentioned, for $f_5$, the proposed algorithm was successfully stabilized by increasing the parameter $k_2$ to the value of 12. However, according to our observations, the increase of $k_2$ did not lead to a completely stable optimization for function $f_2$, such that in both small and large scales, there are 4 failures from the successful optimization (see Table \ref{tab4}). Hence, an increase in the population size is necessary for the stable optimization of $f_2$. Similar to the results in optimization of the function $f_{10}$, IS algorithm had not any success in finding the global minimum of $f_{11}$. However, as indicated in Table \ref{tab4}, despite of GWO and TLBO, IS algorithm shows some stability in finding one of the highly competitive solutions of this problem, i.e., 23 successes of IS against zero success of GWO and TLBO.

%Please make attention that by the regularity of local minimums we mean a function that can be intuitively interpreted as a unimodal function modulated by fine wiggles. As seen in the results for cluster 1, this unimodal background provides an helpful guidance for the algorithms such as GWO and TLBO.

%We should pointing out that a competitive solution do not necessary means a local minimum with a close cost value to the global minimums. Any solution(s) that avoid              This kind of regularity should not be confused with the regularity of multiple global optimum solutions in the function $f_8$.

Finally, the unique feature of the functions in cluster 3 respect to the previous ones is existence of numerous competitive solutions. For example, in two-variable state of the $f_9$ function, the number of competitive solutions is 36 while that was at most 9 among the functions at the cluster 2 (for $f_8$). Further, this number is larger for the function $f_6$, and becomes infinity for $f_{3}$ and $f_{12}$. Both $f_{3}$ and $f_{12}$ have one global optimum solution at the origin. In the $f_{12}$, the flat rings around the origin - with approximately equal optimal cost values as the original point - include infinite number of the competitive solutions. Also, the flat region in $f_3$ contains only existing competitive solutions with the global optimum, which are also infinite in number. As the results in Figure \ref{ss} and Figure \ref{ls} indicate, the proposed algorithm has best performance for the functions in cluster 3, in both small and large scales. At this cluster, most competitive algorithm with the proposed IS algorithm is DE.

Although, in small scale of the problem $f_3$, DE algorithm was defeated by IS just because of one failure (see Table \ref{tab4}), but in 100 dimensions, there is a remarkable difference between the number of successful optimization of IS, i.e., 27, and that of DE algorithm, i.e., 13. DE has a similar instability in optimization of $f_6$ in the large scale. It has 7 unsuccessful optimization according to Table \ref{tab4}, while IS algorithm is completely successful. The \textit{Xin-She Yang 3} function \cite{MJ13} has similar structure as the function $f_6$. For this function, IS algorithm also outperforms other test algorithms according to our observations. The results were omitted for brevity. On the other side, IS algorithm discovers exact optimum solution of the function $f_9$ with minimum number of function evaluations (see Figure \ref{ls}(\textit{i})). Last but not least, the proposed algorithm is the only successful algorithm in approaching the exact optimum solution of $f_{12}$. Although, all comparative algorithms touch the optimal cost value close to zero (see Table \ref{tab3}), but their solutions have large distance from the global optimum solution located in the origin. Figure \ref{ls}(\textit{l}) compares the MSE of estimated solution at minimization of the function $f_{12}$.

\begin{figure}[h!]
 \centering
  \includegraphics[scale=0.32]{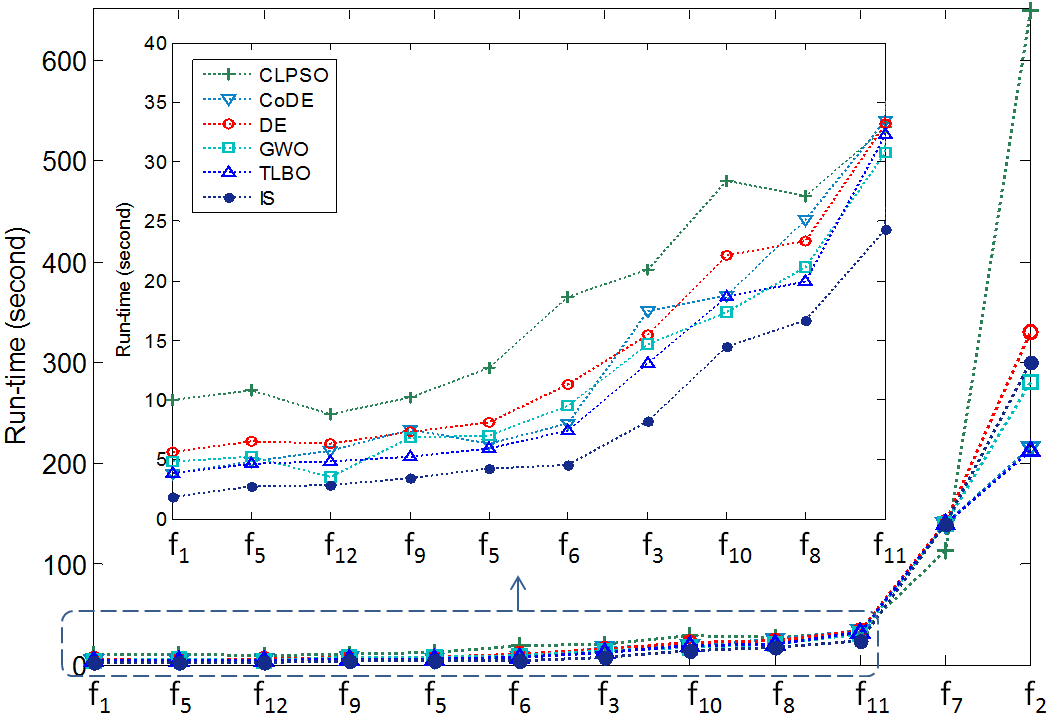}
  \caption{Runtime of test algorithms in optimization of benchmark functions ($f_1-f_9$ are considered in the 100 dimensional case).}
  \label{rt}
\end{figure}

Figure \ref{rt} compares average runtime of the proposed algorithm with that of other test algorithms. At this figure, the benchmark problems were sorted according to their required runtime by IS algorithm. As shown, IS algorithm has smallest runtime in optimization of all benchmark functions except of $f_7$ and $f_2$. In order to have a fair comparison about the complexity of the operations utilized in each algorithm, without taking the complexity of function evaluations into account, we should concentrate at the function $f_1$ as the simplest one for the evaluation. The results for $f_1$ indicate that the complexity of operators of TLBO is 2.0 times more than that of IS algorithm.

\noindent Among the test algorithms, TLBO has simplest operators after IS. Moreover, CLPSO as the most complex one, has a complexity more than 5.3 times. In between, the complexity of DE is exactly 3 times more than that of the proposed algorithm.

\subsection{Sparse Reconstruction}

Sparse reconstruction is generally referred to solving an underdetermined system of linear equations with a prior knowledge of sparsity about the solution. It has a lot of applications such as signal compression \cite{ECG11}, channel estimation \cite{Channel10}, adaptive identification \cite{ACSI}, and spectrum sensing \cite{Spec11}, which specially were developed after compressed sensing theory. According to the compressed sensing \cite{CS06}, a sparse vector $\textbf{x}=[x_1,~x_2,~...,~x_d]^T$ can be recovered from linear measurements of $\textbf{y}= \textbf{A} \textbf{x} + \textbf{n}$, while the number of measurements $m$ is less than the original dimensions of the sparse vector, i.e., $m<d$. A vector is called $k$-sparse, if the number of its non-zero elements are $k$, such that $k\ll d$. At the mentioned linear model, the matrix $\textbf{A}\in \mathbb{R}^{m\times d}$ is known as the measurement matrix and the vector $\textbf{n}\in R^m$ is regarded as the measurement noise. The noise or measurement error is usually modeled by i.i.d. zero-mean Gaussian distribution with an specified variance. A condition for a reliable recovery is holding a degree of randomness by the measurement matrix $\textbf{A}$.  That is satisfied for Gaussian and binary random measurements under a predetermined number of measures \cite{CS06, Uncertainity}.

Reconstruction of the sparse vector $\textbf{x}$ from $\textbf{y}$ and $\textbf{A}$ is an optimization problem. Various optimization models and different algorithms were developed over past decade. Recently, some metaheuristic approaches were proposed for optimization of sparse reconstruction models. Main advantage of metaheuristic approaches is their independency from the properties of the functions used in the optimization model. For example, a preferred model for sparse reconstruction is minimization of the number of non-zero elements ($l_0$ norm) of the solution vector $\mathbf{x}$. Metaheuristic approaches can easily optimize such non-differentiable discontinuous functions. In other approaches, the function should be approximated, e.g., SL0 algorithm in \cite{SL009}. In \cite{nonconvexCS15},\cite{Intel17}, the genetic algorithm was combined with clonal selection and simulated annealing, respectively, to solve the nonconvex $l_0$ minimization problem. Furthermore, another evolutionary algorithm based on a soft-threshold method was earlier developed for the same model \cite{multiob14}. Here, we aimed to optimize the follow single-objective function based on $l_q$ norm:

\begin{equation}\label{lp}
  f_{13}(\textbf{x})= \frac{1}{2} \|\textbf{y}-\textbf{A}\textbf{x}\|_2^p + \lambda \|\textbf{x}\|_q
\end{equation}

\noindent where $\|\textbf{x}\|_q=\sqrt[q]{\sum_{i=1}^{d}x_i^q}$ is the nonconvex operation of $lq$ norm used as a regularization term to promote the sparsity, $\|.\|_2$ indicates the Euclidian norm modeled to minimize the Gaussian measurement errors, such that, its minimization leads to fidelity of the discovered solution to the measurements $\textbf{y}$. Finally, the constant $\lambda$ is regularization coefficient for making a balance between the sparsity and fidelity.

Setting a proper value for $\lambda$ is essential for an accurate reconstruction. Indeed, finding an appropriate value for $\lambda$ in the model (\ref{lp}) with $p=2$ is a tedious task. An advanced approach is separation of the model to two objective functions and utilization of an multiobjective method for finding a good balance between the sparsity-inducing and fidelity functions \cite{multiob14}. At this paper, for sake of simplicity, we target the model (\ref{lp}) with $p=1$. We realized that a valid amount for $\lambda$ is easily approachable by the unit power for fidelity term. In addition, the value of $q$ was set on 0.9. At the following, we first demonstrate efficiency of the proposed algorithm respect to the conventional evolutionary algorithms, and then highlight its possible advantage respect to the state-of-the-art sparse reconstruction algorithms.

First experiment was conducted in two scenarios, both at the case of noiseless measurements with $d=256$ decision variables and $m=128$ measurements. At the first scenario, $20$ nonzero elements of sparse vector $\textbf{x}$ were selected by random and valued by i.i.d. Gaussian distribution with zero mean and unit variance. Also, at this case, the measurement matrix $\textbf{A}$ was a zero-mean Gaussian random matrix with the i.i.d. elements and normalized columns. At the second scenario, the desired sparse vector was binary with $k=20$ nonzero unit elements, distributed by random among the variables of $\textbf{x}$. At this case, the measurement matrix was also binary matrix with equal probability of 0.5 for each 0 and 1 values. The regularization coefficient $\lambda$ was adjusted to 0.1 and 1 for the first and second scenarios, respectively. At both Gaussian and binary scenarios, the parameters were fixed on $Cr=0.2$ and $F=0.4$ for DE, $c_1=2.05$ and $c_2=2.04$ for PSO-cc, and $k_1=20$ and $k_2=2$ for IS, respectively. The number of particles was fixed on 40 for all algorithms.

\begin{table}[h]\small
 \caption{Comparison of distortion and runtime (in millisecond) for sparse reconstruction problems}
\begin{center}
  \begin{tabular}{l  c   c   c  c  c  c  r }
 %\hline
   % use \limits right after \sum of \prod to locate upper/lower bound in right places, e.g. \sum\limits_{}^{}
   % current format is for inside of text
  \hline
  \noalign{\vskip 0.1cm}
          & DE & CoDE & PSO-cc & CLPSO & GWO & TLBO & IS\\

      \hline
  \noalign{\vskip 0.2cm}
\vspace{.02 in}
    Gaussian & 1.3e-3 & 0.132 & 0.574 & 0.609 & 0.397 & 0.676 & 1.2e-3 \\
    \vspace{.05 in}
   % & (4.2e-3) & (9.0e-2) & (0.224) & (0.129) & (0.153) & (0.253) & (3.5e-3) \\
   \vspace{.05 in}
runtime & 25438 &  22142 & 27387 & 33760 & 28794 & 22312 & 19072\\
\vspace{.02 in}
    Binary & 0.010 & 0.304 & 0.780 & 0.825 & 1.048 & 0.691 & 9.10e-4 \\

  \vspace{.05 in}
    runtime & 25402 &  26689 &  31280 & 37916 & 28954 & 22408 & 23203\\

 \hline
  \label{tabo}

\end{tabular}\\
\end{center}
\end{table}

\begin{figure}[h]
 \centering
  \includegraphics[scale=0.32]{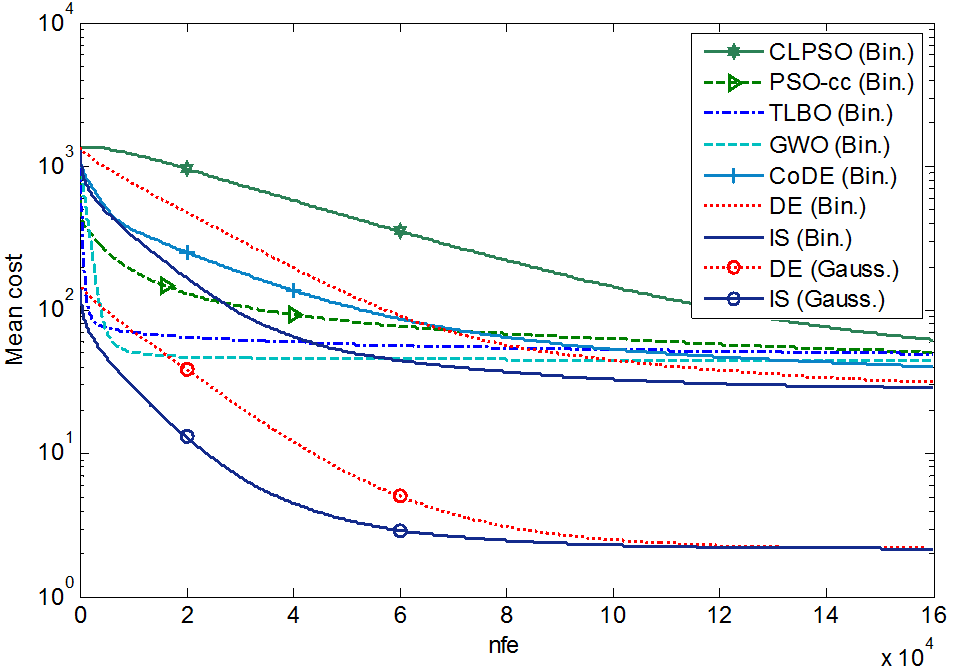}
  \caption{Convergence curves for sparse reconstruction problems}
  \label{CostB}
\end{figure}

Figure \ref{CostB} shows the averaged convergence curve of all test algorithms over 100 trials with different sparse vector and different measurement matrix in each trial. As shown, for the binary scenario, the proposed algorithm (IS) converges to lowest cost value after 160000 function evaluations, while except of DE algorithm, other algorithms are trapped at a local optimum solution (PSO-cc, GWO, and TLBO) or have a slow convergence (CoDE and CLPSO). As can be inferred from Figure \ref{CostB} about the Gaussian scenario, the mentioned number of function evaluations was enough for DE algorithm to capture the same cost value as IS algorithm. Convergence curve of other algorithms were omitted at this scenario, because of their poor performance similar to the binary case. Table \ref{tabo} summarizes the distortion from exact optimal solution in both states. Furthermore, their runtime was included for comparison. As expected, at the Gaussian scenario, the distortion for both DE and IS algorithms are approximately same, while IS algorithm has significantly lower distortion at the binary case. Moreover, at both scenarios, IS algorithm has less runtime than the competitive DE algorithm.

\begin{figure}[h!]
 \centering
  \includegraphics[scale=0.290]{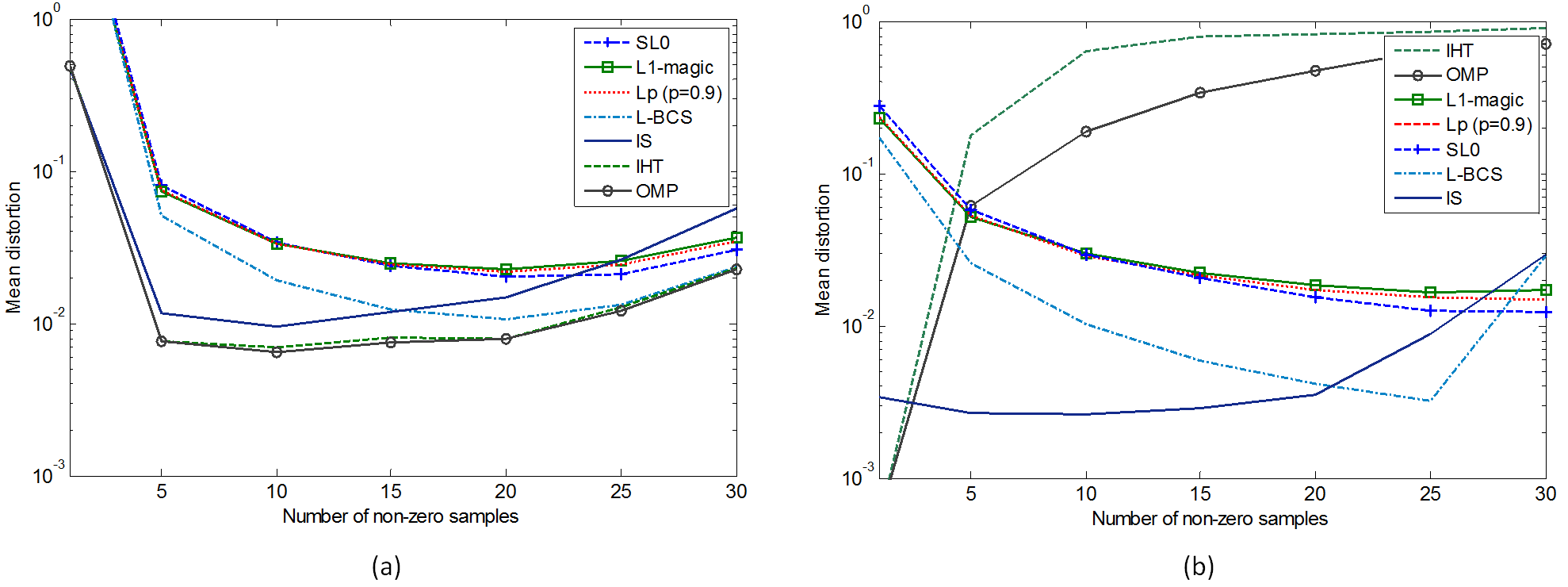}
  \caption{Mean distortion at each level of sparsity for (a) Gaussian and (b) Binary scenarios.}
  \label{disG}
\end{figure}

At the second experiment, IS algorithm was compared with well-known sparse reconstruction algorithms consist of IHT \cite{IHT09} and OMP as the greedy approaches \cite{OMP07}, $l_1$-magic as the conventional interior-point-based optimization method \cite{l1magic}, a Bayesian method with Laplace priors (L-BCS) \cite{lapop}, SL0 that smoothly approximates $l_0$ norm, and $l_q$ algorithm which minimizes an approximated version of the nonconvex function of $l_q$ norm \cite{lp09}. All settings were same as the previous experiment; the dimension of problem was $d=256$, both $l_q$ and IS algorithms were implemented with $q=0.9$, and the parameters of IS algorithm were leaved unchanged. Despite of previous experiment, the measurements were contaminated by noise. Variance of the noise was $1.6\times 10^{-3}$ and $0.04$ for Gaussian and binary scenarios, respectively. The NMSE curves in different number of nonzero elements were plotted in Figure \ref{disG}. As depicted in Figure \ref{disG}(a), for the case of Gaussian sparse vector with Gaussian measurements, the proposed algorithm outperforms all other algorithms except of the greedy ones, in a range of sparsity level, with less than 15 nonzero elements. The better performance of greedy approaches is at expense of a prior knowledge about the number of nonzero elements. In fact, it is not available information in any applications.

On the other side, as depicted in Figure \ref{disG}(b) for binary scenario, IS algorithm has better performance than all other algorithms when the optimal sparse solution has more than 2 and less than 20 nonzero elements. Despite of Gaussian scenario, for binary signals, OMP and IHT algori-

\noindent thms have unstable performance, since, identification of nonzero elements with the same values is generally hard for greedy approaches. The price for such outstanding performance of an evolutionary algorithm is a large runtime, even at the several order of magnitudes more than the sparse reconstruction algorithms. The worst runtime among the dedicated algorithms was approximately around 0.7 second for the $l_q$ algorithm. Parallel computing is main technique for reduction of runtime, with open problems in order to approach a real-time implementation for the population-based approaches \cite{RS17}.

\subsection{Antenna Selection in Massive MIMO}

Multiuser multiple-input-multiple-output (MU-MIMO) communication system is fundamental technology in wireless networks. The system is modeled by linear equations of $\textbf{y}=\textbf{DH}\textbf{s}+\textbf{n}$, where, $\textbf{DH}\in \mathbb{C}^{m\times d}$ is the channel matrix between $d$ antennas at the base station (BS) and $m$ single-antenna mobile stations (MSs). It contains large-scale fading coefficients (path loss and shadowing) as the elements of diagonal matrix of $\textbf{D}\in \mathbb{R}^{m\times m}$, and small-scale fading coefficients modeled by matrix $\textbf{H}\in \mathbb{C}^{m\times d}$ with i.i.d. complex random variables driven from Gaussian distribution with mean of zero, and unit variance. The elements of vector $\textbf{s}\in \mathbb{C}^{d\times 1}$ are transmitted samples from each antenna of BS (at downlink), $\textbf{y}\in \mathbb{C}^{m}$ consists of received samples at the MSs, and $\textbf{n}\in \mathbb{C}^{m}$ is the noise, modeled at the receiver side \cite{DT05}. Recent developments in MU-MIMO is based on large number of antennas at BS that support several MSs. As the number of BS antennas scales up, the interference among MSs vanishes and simple linear precoding/combinig methods provide near-capacity performance \cite{MaMi14}. However, deployment of large number of radio frequency (RF) chains, one per each antenna, leads to high cost and energy consumption. Antenna selection is a way to tackle this issues \cite{XG15}.

The number of RF-chains can be reduced by selecting an optimal subset of antennas and deactivating rest of them. Two well-known selection criterions are based on maximization of achievable sum throughput and maximization of minimum singular value of the selected channel matrix. According to \cite{RH01}, at the case of utilization of zero-forcing (ZF) precoder, the minimum received signal-to-noise ratio (SNR) among MSs is lower bounded by an scale of squared minimum singular value (MSV) of the selected channels for transmission. Maximization of this parameter leads to maximization of minimum received SNR among the MSs, or equivalently minimization of bit error rate. Hence, in the network terminology, a minimum quality of service (QoS) can be guaranteed by the MSV criteria. Fitness function for maximization of MSV is formulated as the following combinatorial model:%\footnote{For convenience on interpreting the fitness values, the MSV is maximized instead of its squared version. It has not qualitative effect on the results.}\footnote{The model is not convex nor concave.}

\begin{equation}\label{msv}
  f_{14}(\textbf{x})= \min {\sigma (\textbf{HX})} ~~~ s.t. ~ \textbf{x}\in \{0,1\}, ~ \|\textbf{x}\|_0=k
\end{equation}

\noindent where binary decision variables of vector $\textbf{x}$ are the elements of $d\times d$ diagonal matrix $\textbf{X}$, i.e., 1 for selected antennas and 0 for deactivated antennas. The operator $\sigma(.)$, calculates singular values of the selected channels, and cardinality of vector $\textbf{x}$ is constrained to the predefined number of active antennas, $k$. Without any manipulation in the algorithm, $f_{14}(\textbf{x})$ was replaced by $-f_{14}(\textbf{x})$ for the aim of maximization. In addition, the algorithm was implemented at the continuous domain. Initial solutions were produced randomly at the range of $[0,1]$, with uniform distribution. The only modification was mapping of the continuous decision variables to the binary digits, before function evaluation at the step 3 of Algorithm 1. The mapping was simply accomplished through replacement of $k$-largest decision variables by integer value of 1 (indicating the selected antennas), and discarding other variables to zero.

Maximization of MSV has also applications in the other engineering problems \cite{BB08}. In the context of wireless communication, speed of optimization is main challenge, specially in large-scale antennas. A branch and bound (BAB) method was developed for maximization of MSV in \cite{BB08}. The BAB-based algorithms approach the exact optimum solution with a reduced complexity respect to the exhaustive search (ES). The BAB algorithm in \cite{BB08} was considered for large-scale antenna selection in \cite{BB15}. According to their results, the algorithm provides 4 order of magnitude reduction at the required number of function evaluations, in comparison with ES. However, this reduction is not sufficiently enough for the large scales. In addition, their algorithm is developed for selection of a square matrix. On the other words, the number of selected antennas is always equal to the number of MSs. Metaheuristic algorithms are efficient alternative for approximation of optimal solution with significantly less complexity, as well as flexility on the number of selected antennas.

Performance of the test algorithms were not competitive with the proposed algorithm, expect of DE algorithm. We focus to compare the results with DE algorithm, and a recently developed binary optimizer for large-scale antenna selection, based on genetic algorithm (GA) \cite{Makki17}. In all of the simulations, the parameters were set on $k_1=39$ and $k_2=10$ for IS algorithm, and $Cr=0.2$ and $F=0.1$ for DE algorithm, both with the population size of 40. According to our observations, the utilized parameter values in the GA-based algorithm for maximizing the sum throughput \cite{Makki17}, were also efficient for MSV maximization. Hence, its population size was set on 10, with 5 of them as the mutation vectors on the best solution. The number of BS antennas and number of single-antenna MSs were fixed on $d=128$ and $m=10$, respectively. The results were averaged out over 400 channel realizations with normalised channel vectors of each MS \cite{XG15},\cite{XG15Jul}. At each trial, MSs were uniformly distributed on a hexagonal cell with radius 500m around BS. The path loss exponent was set on 3.8 and mean of shadowing attenuation - modeled by lognormal distribution - was fixed on 8dB.

\begin{figure}[h]
 \centering
  \includegraphics[scale=0.32]{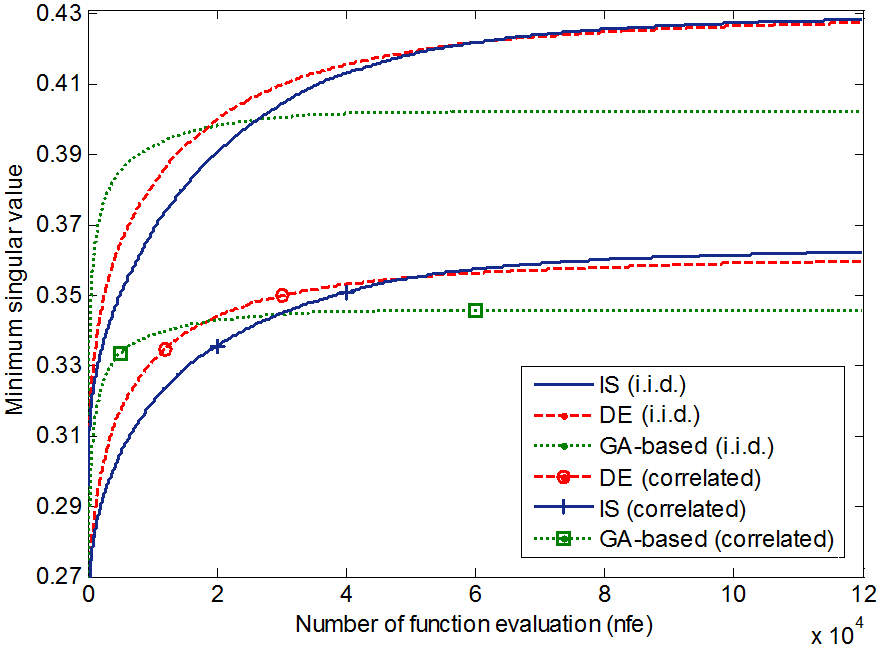}
  \caption{Convergence curves to maximum MSV for i.i.d. and correlated channels.}
  \label{TAS}
\end{figure}

Figure \ref{TAS} shows the convergence curves at the cases of i.i.d. and correlated channels. The correlation among BS antennas changes the i.i.d. small-scale fading channel $\mathbf{H}$ to the correlated channel $\mathbf{H}_c=\mathbf{H}\sqrt{\mathbf{R}}$, where, $\mathbf{R}\in \mathbb{R}^{d\times d}$ is the exponential correlation matrix, defined by $R_{i,j}=\alpha^{|i-j|}$ for its $(i,j)-$th element. At this experiment, correlation coefficient $\alpha$ was set to $0.3$, and the number of selected antennas was $k=30$. As shown in the figure, IS algorithm performs slightly better than DE and outperform GA-based algorithm, in both i.i.d. and correlated channels. It is worth mentioning that, with this parameters of the system, the exhaustive search requires more than $1.5\times10^{29}$ function evaluations, equal to the number of possible combinations, i.e., $\binom{d}{k}=\frac{d!}{(d-k)!k!}$, while, the proposed algorithm converges in only $1.2\times10^5$ function evaluations. It means 24 order of magnitude reduction in the complexity, which is significantly less than the four order reduction by bidirectional BAB algorithm \cite{BB15}. The experiment with i.i.d. channel was repeated by imperfect channel state information (CSI). The perfect CSI is modeled by $\mathbf{H}=\beta \hat{\mathbf{H}}+\sqrt{1-\beta ^2}\tilde{\mathbf{H}}$, where, $\hat{\mathbf{H}}$ is measured available information and $\tilde{\mathbf{H}}$ is the unknown error with the same i.i.d. complex Gaussian distribution as the $\hat{\mathbf{H}}$ matrix. The scaler $\beta \in [0,1]$ determines the accuracy of available CSI. Similar to the results in \cite{Makki17}, here, at the case of MSV maximization, the performance of antenna selection algorithms also degrades significantly as the quality of CSI decreases in small values of $\beta$. As instance, for $\beta = 0.9$, the minimum singular value of the selected channel by IS, DE, and GA-based algorithms are reduced to 0.345, 0.347, and 0.326, respectively.  The results indicate similar sensitivity of IS and DE algorithms to this deficiency.

\begin{figure}[h]
 \centering
  \includegraphics[scale=0.32]{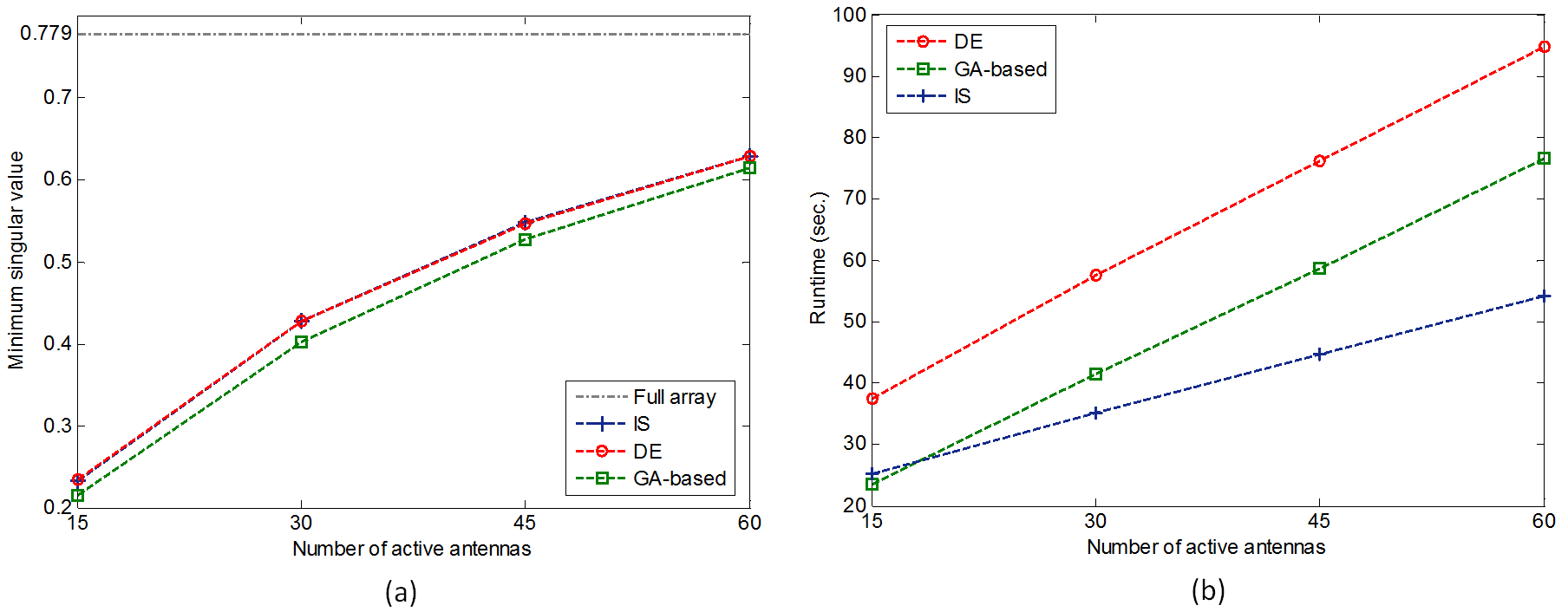}
  \caption{Results of selection with 15, 30, 45, and 60 active antennas; (a) minimum singular value of selected channels, (b) runtime.}
  \label{Speed}
\end{figure}

Finally, with the same parameters, the algorithms were evaluated on different number of selected antennas. Figure \ref{Speed}(a) shows the achieved MSVs at four levels of active antennas, i.e., 15, 30, 45, and 60. All algorithms were stopped after 120000 function evaluations. The channels were uncorrelated with an available perfect CSI. Approximately, same performance of IS and DE algorithms indicates their similar sensitivity to their parameters. The results were also compared with the MSV at the case of full array without antenna selection. The obtained MSVs by IS and DE algorithms for 60 optimal subset of antennas are approximately $\%80$ of the MSV with full use of 128 antennas without any selection. Main advantage of the proposed algorithm is its computational speed. Figure \ref{Speed}(b) illustrates average runtime in the different number of active antennas ($k$). As $k$ increases, the runtime for DE and GA-based algorithms increases approximately at the same rate, while this rate for IS algorithm is roughly half of them. We have attention that for $k=30$, the runtime can be reduced by decreasing the $nfe$, without significant loss in performance. Moreover, at low number of BS antennas, the running times are significantly less, with similar respective performances illustrated for 128 dimensional antenna array. For example, with $d=64$ BS antennas, $m=6$ MSs, and $k=16$ active antennas, the required $nfe$ for convergence is 40000, that takes on average 9.2 and 5.8 seconds running time by DE and IS algorithms, respectively.

\section{Discussion} \label{discussion}

Grouping method and update rule have pivotal influence in the functioning of the metaheuristic algorithms. As shown by numerical results on benchmark problems, the proposed dialectical grouping has significant performance in optimization of specific functions with large number of competitive solutions. However, finding a common feature for all possible problems in which a metaheuristc algorithm has superior performance is really hard. The proposed interactions lead to a simple and delicate step-size mechanism. Numerous experiments confirm convergence of IS algorithm under the suggested mechanism. A mathematical proof of optimality of these step-sizes is a challenging task. In general, the analysis of metaheuristic algorithms is a challenging problem because of existence of various sources of random operations. However, we are hopeful that our proposed deterministic interactions for speculation (as a counterpart for conventional mutation operator) would simplify the analysis of IS algorithm. The two parameters of IS algorithm were easily tuned because of their integer identities and low number of possible pairs that influence on the performance. Nevertheless, an adaptive scheme for adjustment of parameters during the optimization process and its optimality remains as an open problem. As the future research directions, extension of the algorithm to the multiobjective scenarios, and applications in other engineering and scientific problems is of interest.

%We also simulated the antenna selection based on the sum rate maximization. According to our observations, IS and DE algorithms show same performance as the GA-based algorithm, however, with larger number of required function evaluations. The results are omitted for brevity. As an inference, our defined mutation operator based on the idealistic dialectic has less sensitivity to the considered optimization models, than the neighborhood-based mutations in the GA-based algorithm \cite{Makki17}.

%The GA-based algorithm had shown optimal performance in large-scale antenna selection based on maximum sum rate \cite{Makki17}.

\section{Conclusion} \label{conclusion}

Philosophical paradigm of thesis-antithesis-synthesis in dialectical thinking modes promises an efficient search approach. Inspired by speculative and practical thinking modes, we developed a new population-based optimization approach. Speculative thinking - assigned to high quality solutions - was modeled in a way that boosts exploration capability of the proposed algorithm. At this thinking mode, each particle/thinker looks for another one in the community who has a solution (thesis) in largest distance but with similar quality (idealistic antithesis). In contradiction, practical thinking - assigned to low quality solutions - exploits efficiency of the best solution or its idealistic antithesis by selecting one of them which is in smaller distance (materialistic antithesis). Detected antitheses were used as a reference point for reformation (update) of the solutions/theses. Uniformly distributed step-sizes with a negligible bias toward the antithesis, were utilized for explorative speculations, and a biased Gaussian distribution was used for step-sizes of exploitive practices. Results indicate efficiency of the proposed optimization scheme by low-complexity operators.

\bibliographystyle{ieeetr}
{\footnotesize
\bibliography{IS}}

%\bibliography{IS}

\end{document}